\pdfoutput=1
\documentclass[nohyperref]{article}
\usepackage{amsmath,amsfonts,bm}

\def\eqref#1{equation~\ref{#1}}
\def\1{\bm{1}}

\DeclareMathAlphabet{\mathsfit}{\encodingdefault}{\sfdefault}{m}{sl}
\SetMathAlphabet{\mathsfit}{bold}{\encodingdefault}{\sfdefault}{bx}{n}

\DeclareMathOperator*{\argmax}{arg\,max}

\usepackage{enumitem}
\newcommand{\bx}{\mathbf{x}}
\newcommand{\bw}{\mathbf{w}}

\usepackage{setspace}
\usepackage{microtype}
\usepackage{graphicx}
\usepackage{subfigure}
\usepackage{booktabs} % for professional tables

\usepackage{hyperref}

\usepackage[accepted]{icml2022}

\usepackage{amsmath}
\usepackage{amssymb}
\usepackage{mathtools}
\usepackage{amsthm}

\newcommand\defeq{\stackrel{\mathclap{\tiny\mbox{def}}}{=}}

\usepackage{bbm}
\usepackage{dsfont}
\usepackage{caption}
\usepackage{graphicx}

\usepackage{tikz}

\usepackage[capitalize,noabbrev]{cleveref}

\theoremstyle{plain}
\newtheorem{theorem}{Theorem}[section]
\newtheorem{proposition}[theorem]{Proposition}

\theoremstyle{definition}
\newtheorem{definition}[theorem]{Definition}

\theoremstyle{remark}

\usepackage[disable,textsize=tiny]{todonotes}
\icmltitlerunning{Robust Ranking Explanations}

\begin{document}

\twocolumn[
\icmltitle{Robust Ranking Explanations}



\icmlsetsymbol{equal}{*}

\begin{icmlauthorlist}
\icmlauthor{Chao Chen}{lu}
\icmlauthor{Chenghua Guo}{bupt}
\icmlauthor{Guixiang Ma}{uic}
\icmlauthor{Ming Zeng}{cmu}
\icmlauthor{Xi Zhang}{bupt}
\icmlauthor{Sihong Xie}{lu}
\end{icmlauthorlist}

\icmlaffiliation{lu}{Lehigh University, PA, USA}
\icmlaffiliation{bupt}{Beijing University of Posts and Telecommunications, Beijing, China}
\icmlaffiliation{uic}{University of Illinois Chicago, IL, USA}
\icmlaffiliation{cmu}{Carnegie Mellon University, PA, USA}

\icmlcorrespondingauthor{Sihong Xie}{xiesihong1@gmail.com}

\icmlkeywords{Explainable Machine Learning, Robustness}

\vskip 0.3in
]



\printAffiliationsAndNotice{}  % leave blank if no need to mention equal contribution

\begin{abstract}
Robust explanations of machine learning models are critical to establish human trust in the models.
Due to limited cognition capability,
most humans can only interpret the top few salient features. 
It is critical to make top salient features robust to adversarial attacks,
especially those against the more vulnerable gradient-based explanations.
Existing defense 
measures robustness using $\ell_p$-norms,
which have weaker protection power. 
% Gradient-based explanation is the cornerstone of explainable deep networks, but it has been shown to be vulnerable to adversarial attacks.
% % and many works propose defense strategies for better robustness.
% However,
% existing works
% % on either attack or defense side,
% measure the explanation robustness based on $\ell_p$-norm,
% which can be counter-intuitive to humans, 
% who only pay attention to the top few salient features.
We define explanation thickness for measuring salient features ranking stability,
% We then present a new practical adversarial attacking goal for manipulating explanation rankings.
% To mitigate the ranking-based attacks 
and derive tractable surrogate bounds of the thickness to design the \textit{R2ET} algorithm to efficiently maximize the thickness and anchor top salient features.
Theoretically, we prove a connection between R2ET and adversarial training.
% ; we further formulate a constrained multi-objective optimization problem and an attacking algorithm to prove a \textit{global} convergence rate, linking thickness maximization to ranking stability.
Experiments with a wide spectrum of network architectures and data modalities,
including brain networks, 
demonstrate that R2ET attains higher explanation robustness under stealthy attacks while retaining accuracy.
% We also experimentally check the coherence between the thickness and the explanation robustness.
% even more robust than widely accepted Hessian-based curvature smoothing approaches.
\end{abstract}

\section{Introduction}
\label{sec:intro}

\begin{figure}
    \centering
% \begin{minipage}{.22\textwidth}
%     \includegraphics[width=\textwidth]{figs/pipeline.pdf}
% \end{minipage}%
% \enspace
\begin{minipage}{.22\textwidth}
    \includegraphics[width=\textwidth]{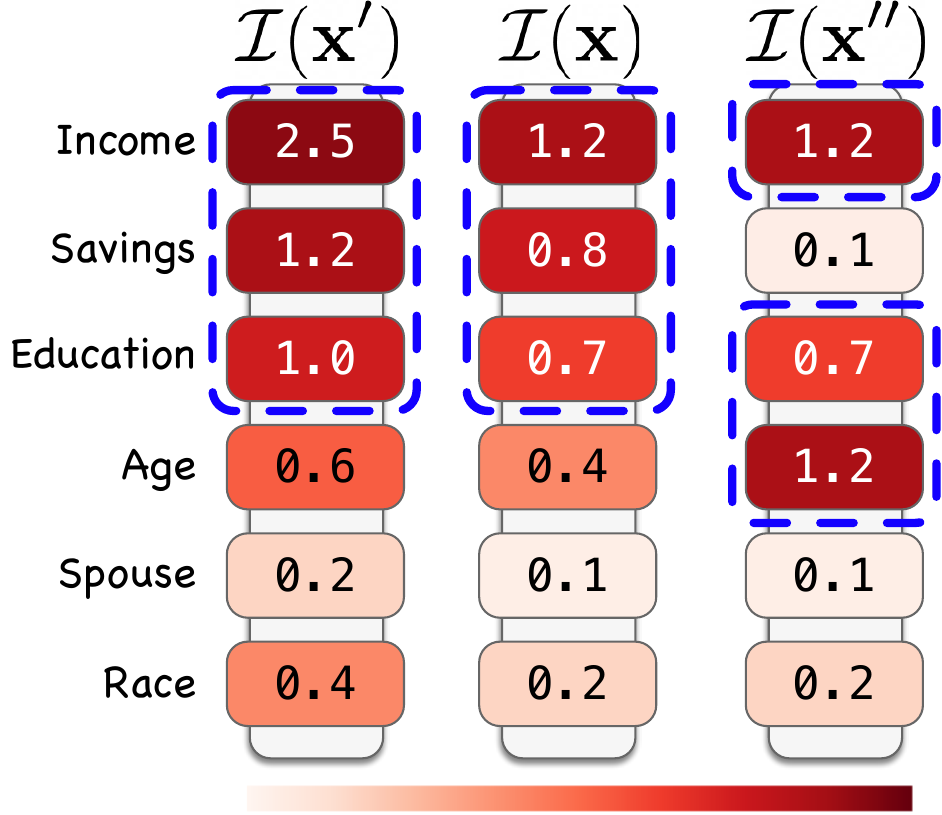}
\end{minipage}%
% \qquad
\enspace
\begin{minipage}{.25\textwidth}
    \includegraphics[width=1.0\textwidth]{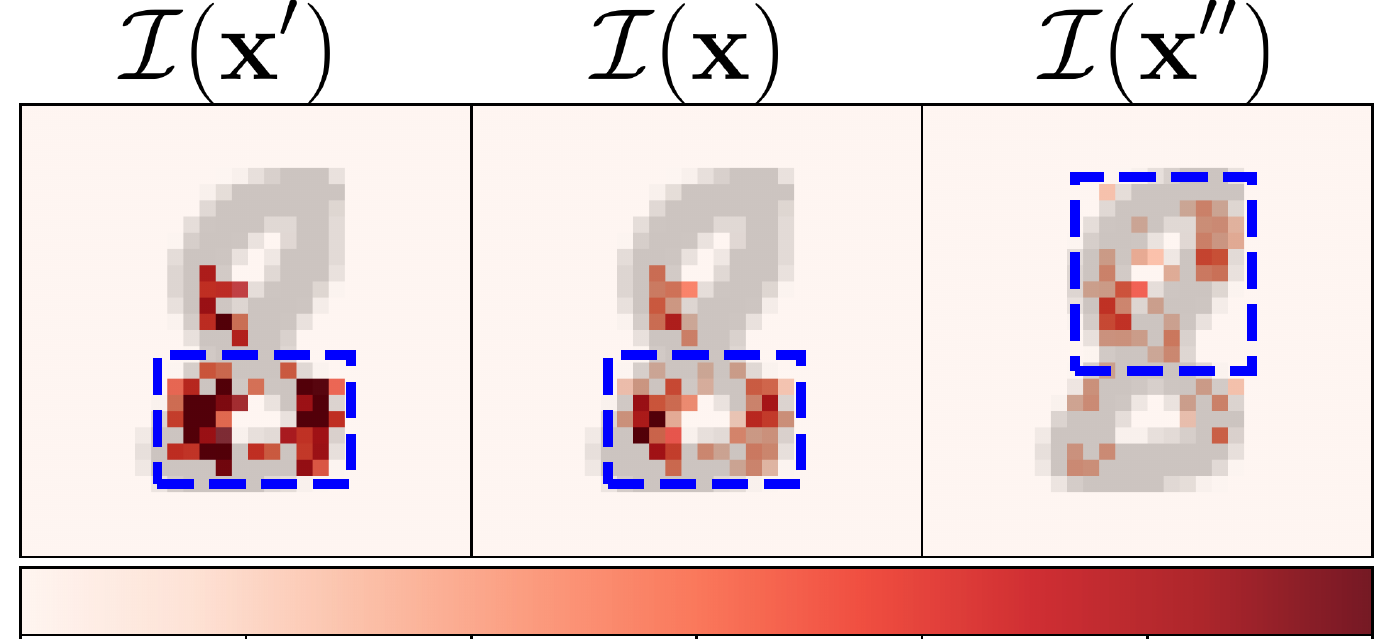}
\end{minipage}    
    \caption{\small A smaller $\ell_p$ distance between saliency maps does not imply similar top salient features.
    $\mathbf{x}$ is the original input,
    and $\mathbf{x}^\prime$ and $\mathbf{x}^{\prime\prime}$ are two perturbed inputs.
    Saliency map (explanation), denoted by $\mathcal{I}(\cdot)$, is a function of the input.
    The saturation of the red color indicates feature saliency,
    and the blue dashed boxes highlight the top salient features/regions.
    \textbf{Left}:
    $\|\mathcal{I}(\mathbf{x}) - \mathcal{I}(\mathbf{x}^\prime)\|_2 
    = 2.4 > 1.1 = 
    \|\mathcal{I}(\mathbf{x}) - \mathcal{I}(\mathbf{x}^{\prime\prime})\|_2$.
    However, $\mathcal{I}(\mathbf{x}^{\prime})$ and $\mathcal{I}(\mathbf{x})$ have
    the same top-$3$ salient features.
    \textbf{Right}:
    $\| \mathcal{I}(\mathbf{x}) -  \mathcal{I}(\mathbf{x}^\prime)\|_2 = 0.10> 0.07=\|\mathcal{I}(\mathbf{x})-\mathcal{I}(\mathbf{x}^{\prime\prime})\|_2$,
    but the top-$50$ salient features from 
    $\mathcal{I}(\mathbf{x})$ and $\mathcal{I}(\mathbf{x}^\prime)$ have a 92$\%$ of overlap, and only
    % the intersection of
    36$\%$ between  
    $\mathcal{I}(\mathbf{x})$ and
    $\mathcal{I}(\mathbf{x}^{\prime\prime})$.
    % \textcolor{red}{The $\mathcal{I}(x)$ symbols should be made more smooth. Try to generate the symbols using LatexIT and then copy the images to the Omnigraffle image.}
    }
    \label{fig:intro_example}
    % \vspace{-0.1in}
\end{figure}

Deep neural networks (DNNs) have proven their strengths in many real-world applications, 
including 
financial~\cite{wang2020using}, 
image retrieval~\cite{zhou2020adversarial}, 
and biomedical research~\cite{hudson2000neural}.
% However, the lack of human trust prevents wider applications of DNNs where stakes are high.
The explainability of DNNs is a fundamental requirement for establishing  humans' trust and is key to further deployments in high-stake applications~\cite{goodman2017european,pu2006trust}.
% Explaining a model's predictions can help establish trust~\cite{goodman2017european,pu2006trust}.
% We observe that the top salient features are more relevant to 
As human cognitive capability is limited~\cite{saaty2003magic},
an explanation typically attributes a prediction to a few salient features of the input data (see Fig. \ref{fig:intro_example}).
% and can only focus on the top salient features to drive their decision-making~\cite{tsetsos2012salience}.
Among all explanation methods, 
saliency maps based on model gradients with respect to input data are widely adopted due to their inexpensive computation and intuitive interpretation \cite{nielsen2022robust}.
% , which play the most important roles for explanations.

% Of all explanation methods, Saliency maps are commonly used due to their inexpensive computation and intuitive interpretation \cite{nielsen2022robust}.
% % that assigns each input feature an important score based on gradients of the model input with respect to the corresponding feature. 
% % Unfortunately,
% Existing works \cite{dombrowski2019explanations,dombrowski2021towards,ghorbani2019interpretation}
% show that 
% the gradients may not be robust and can be manipulated
% without changing the model input and output to the human-perceptible extent
% leading to inaccurate explanations and mistrust of the model.

\begin{figure}
    \centering
    \includegraphics[width=.45\textwidth]{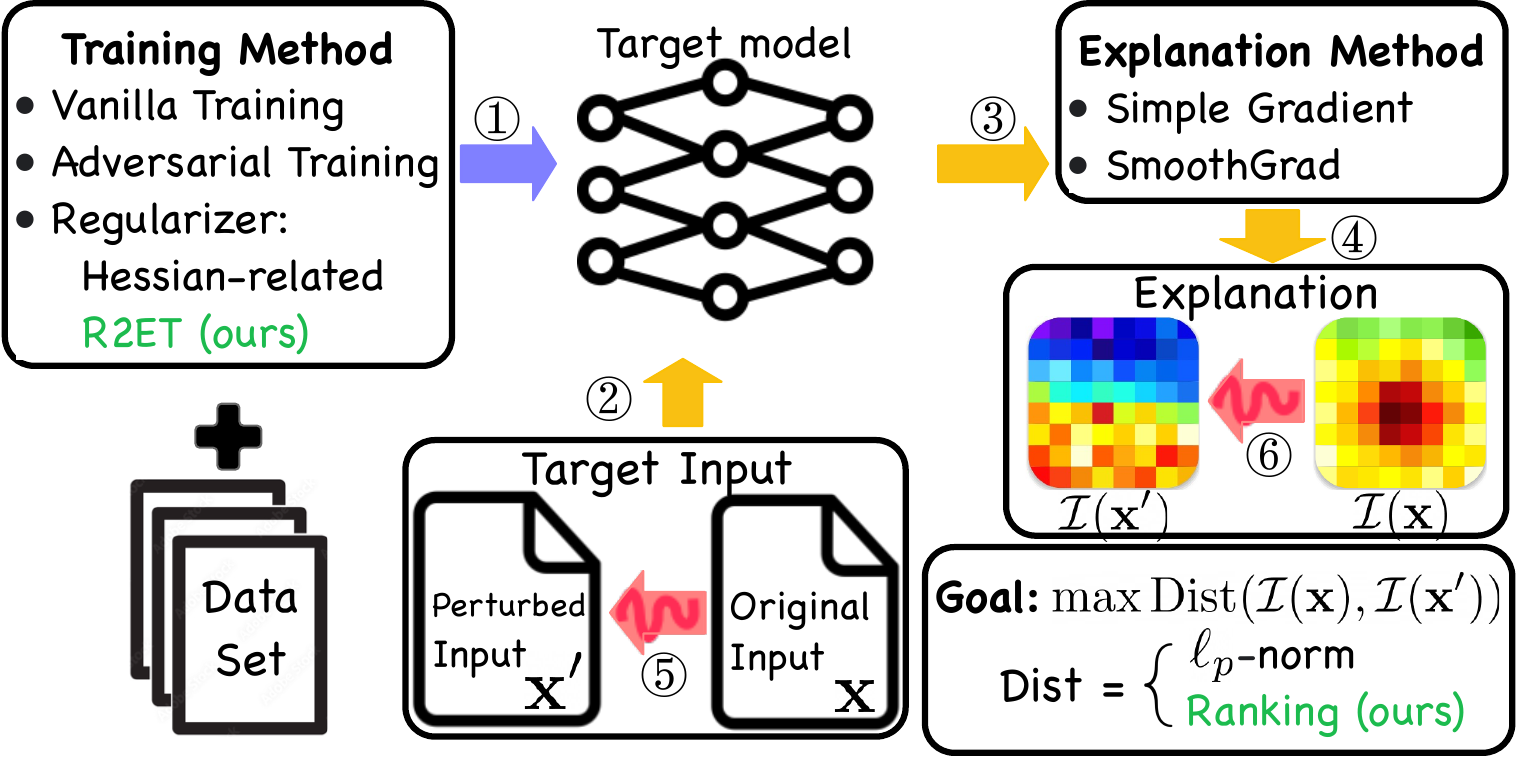}
    \caption{\small 
    \textbf{Blue} (\textcircled{\raisebox{-0.9pt}{1}}): Model training. 
    \textbf{Yellow} (\textcircled{\raisebox{-0.9pt}{2}}-\textcircled{\raisebox{-0.9pt}{4}}): Explanation generation for a target input.
    \textbf{Red} (\textcircled{\raisebox{-0.9pt}{5}}-\textcircled{\raisebox{-0.9pt}{6}}): Adversarial attacks against the explanation by manipulating the input.
    % \textbf{Grey} arrows: Backward propagation for conducting attacks.
    % \textcolor{red}{The dist function should connect the two $\mathcal{I}$ functions}
    }
    \label{fig:pipeline}
    % \vspace{-0.1in}
\end{figure}

% that assigns each input feature an important score based on gradients of the model input with respect to the corresponding feature. 
% Unfortunately,
Existing works \cite{dombrowski2019explanations,ghorbani2019interpretation,heo2019fooling}
show that 
the gradients can be manipulated with unnoticeable changes in the input.
% leading to inaccurate explanations and mistrust of the model
% (shown in Fig. \ref{fig:pipeline}).
They measure the explanation robustness using a certain $\ell_p$ norm,
leading to 
% Existing works on manipulating explanations (shown by red arrows in the left of Figure \ref{fig:intro_example})
% \cite{heo2019fooling,dombrowski2019explanations,dombrowski2021towards,ghorbani2019interpretation}
% ,wang2020smoothed}
% measure the robustness by $\ell_p$ norm \cite{heo2019fooling,ghorbani2019interpretation}.
the idea of minimizing the $\ell_p$ norm of the Hessian of the model against the input for robust explanations ~\cite{dombrowski2021towards,wang2020smoothed}.
% However,
% % and care about the \textit{magnitude} of importance score for each feature independently.
% % We refer MSE attack to maximize the $\ell_2$-norm between the original and perturbed explanations as goal.
% we observe that the top salient features are more relevant to humans who consume the explanations: 
% % The changes of the top few important features can have a deeper impact on human's perceptions.
% humans have very limited attention capacity~\cite{saaty2003magic} and can only focus on the top salient features to drive their decision-making~\cite{tsetsos2012salience}.
However, as demonstrated in Fig.~\ref{fig:intro_example},
a perturbed explanation with a small $\ell_p$ distance to the original one can have rather different top salient features,
since the $\ell_p$ norm considers the importance of \textit{all} features equally.
Such inconsistency between the $\ell_p$ metric and the modus operandi of human perception can lead to mistrust of the model and the associated explanations.
% In other words,
% the explanation robustness measured $\ell_p$ norm can be counter-intuitive to humans,
Alternatively,
we will measure the stability of the rankings of the top salient features.
Prior work on ranking robustness can be found in information retrieval \cite{zhou2020adversarial,zhou2021practical}, though they are not applicable to stability in explainable ML,
in terms of vulnerabilities, attacking objectives, theoretical analysis, and computations
(see Sec. \ref{sec:related_work}).

\textbf{Contributions.}
We center our contributions around a novel metric called ``ranking explanation thickness'' that precisely measures
% and optimize 
the robustness of the top salient features.
% without expensive adversarial training.
(1)
Theoretically,
we derive surrogate bounds of ranking explanation thickness for more efficient optimization and 
to reveal a fundamental limit of using Hessian norm for explanation robustness.
We also disclose the equivalence between ranking explanation thickness and Adversarial Training in Eq. (\ref{eq:obj_at}).
% Sec.\ref{sec:analysis} justifies maximizing thickness for explanation robustness. 
(2)
Algorithmically,
based on the theoretical analysis,
we propose an efficient training method, R2ET (see Fig. \ref{fig:pipeline}), to attain the desired robustness without costly adversarial training.
R2ET optimizes the surrogate bounds to overcome the limitation of existing Hessian-based curvature smoothing approaches.
(3)
Experimentally, we conduct both $\ell_2$-norm attack and a ranking-based attack (see Fig. \ref{fig:pipeline}) to demonstrate: 
% \textit{i}) the limit of prior attacking and defense algorithms based on $\ell_p$ norm and Hessian norm, respectively; 
\textit{i}) prior $\ell_p$-norm attack cannot manipulate as efficiently as ranking-based attacks,
and Hessian-related defense algorithms do not result in robust rankings;
\textit{ii}) the high correlation between explanation thickness and robustness;
\textit{iii}) the generality and wide applications of R2ET to different neural network architectures 
% (single-input DNNs and dual-input Siamese Networks \cite{ma2019deep}) 
and 
% datasets (tabular, image, and graph data).
various data modalities,
including two brain network datasets.
\section{Preliminaries}
\label{sec:preliminary}
\noindent\textbf{Saliency map explanations.}
% Consider a classification problem on the domain $\mathcal{X}$.
Let a classification model with parameters $\bw$ be $f(\mathbf{x}, \bw):\mathcal{X} \to [0,1]^C$, and $f(\mathbf{x}, \bw)_c$ is the probability of class $c$ for the input $\mathbf{x} \in \mathcal{X}$.
% and $\bw$ the model parameters.
We omit $\bw$ for brevity in the sequel as $\bw$ is fixed.
% We restrict to the class $\argmax_c f(\mathbf{x})_c$ if not specified.
% Existing works [cite] consider gradients as explanations 
Gradient-based methods \cite{adebayo2018sanity,baldassarre2019explainability} 
% suppose that a larger gradient magnitude indicates more sensitivity of $f(\mathbf{x})_c$ to the corresponding features,
% and thus 
explain $f(\mathbf{x})_c$ by the top features with the largest magnitudes.
Formally,
a saliency map explanation is defined as $\mathcal{I}(\mathbf{x},c;f) = \nabla_{\mathbf{x}} f(\mathbf{x})_c$.
% with respect to these components of input.
% For example, 
% the prediction of a model is more likely to change due to the perturbations on the components with large magnitude than those with small magnitude.
% Thus, 
% gradient-based methods \cite{baldassarre2019explainability,schnake2020higher} are widely used to explain the prediction of a target sample $\mathbf{x}_i$ from a single-input classification model $f(\mathbf{x})$.
% the explanation of $f(\mathbf{x})$ is
% \begin{equation}
%     \label{eq:explain_general}
%     \mathcal{I}(\mathbf{x}) 
%     = \nabla_{\mathbf{x}} f(\mathbf{x}),
% \end{equation}
% where $\ell$ is a classification loss used in the training phase, such as cross entropy.
% A variant is to take the element-wise multiplication
% $\mathcal{I}(\mathbf{x}; f) = \mathbf{x} \circ \nabla_{\mathbf{x}} f(\mathbf{x})$ as the explanation \cite{adebayo2018sanity}.
Since the model $f$ is fixed for explanations and we fix $c$ to the \textit{predicted} class,
% and all classes can be treated the same way,
we use $\mathcal{I}(\mathbf{x})$ to denote $\mathcal{I}(\mathbf{x},c;f)$,
and use $\mathcal{I}(\mathbf{x})_i$ to denote the $i$-th feature's importance score.
\noindent\textbf{Threat model.}
The adversary solves the following problem
% the goal 
to find the optimal perturbation $\delta^\ast$  
% whose $\ell_2$ norm is not larger than a given budget $\epsilon$
to distort the explanations without changing the predictions
% \cite{ghorbani2019interpretation,rony2020augmented}:
\cite{dombrowski2021towards}:
\begin{equation}
    \label{eq:attack_general}
	%\setlength{\jot}{3pt}
% 	\begin{gathered}[t]
% 	\begin{split}
    \begin{aligned}[t]
% 	\max_{\delta} 
    \max_{\delta: \|\delta\|_2 \leq \epsilon}
	~~ & \textnormal{Dist}
    (\mathcal{I}(\mathbf{x}), \mathcal{I}(\mathbf{x}+\delta)) \\
    \mbox{s.t.} ~~ &
    % \textnormal{s.t.} &
    \argmax_c f(\mathbf{x})_c = 
    \argmax_c f(\mathbf{x}+\delta)_c,
    % f(\mathbf{x}) \approx f(\mathbf{x}+\delta),
    % \quad
    % \|\delta\|_2 \leq \epsilon.
% 	\min_{\bt_u \in \Theta_u \subseteq \mathbb{R}^n, \, \bt_\ell \in \mathbb{R}^m} ~~ & f_u(\bt_u,\bt_\ell) \\
% 	\mbox{s.t.}~~ & \bt_\ell \in \argmin ~~ f_\ell(\bt_u,\bt_\ell),\\
    \end{aligned}
% 	\end{split}
% 	\end{gathered}
	\end{equation}
% \textcolor{red}{Also explain $\delta^\ast$}.
% $\delta^\ast$ is the optimal perturbation and $\mathbf{x}^\prime=\mathbf{x}+\delta$ is the corresponding adversarial sample.
$\delta$ is the perturbation whose $\ell_2$-norm is not larger than a given budget $\epsilon$.
$\textnormal{Dist}(\cdot,\cdot)$
% usually the $\ell_p$-norm 
% % \cite{dombrowski2019explanations,dombrowski2021towards,wang2020smoothed},
% \cite{dombrowski2019explanations},
evaluates how different the two explanations are.
% In this paper, 
E.g., $\textnormal{Dist}(\cdot,\cdot)$ is inversely related to the correlation between two rankings.
% In terms of adversarial attacks against rankings,
% $\textnormal{Dist}(\cdot,\cdot)$ is designed as the differentiable proxy ranking loss \cite{wang2020transferable,zhou2020adversarial,zhou2021practical},
% such as the summation of the distance between flipped pairs of features in the perturbed rankings.
% % to mess up the original rankings.
% $\|\cdot\|_2$ is the $\ell_2$ norm.
% The constraint 
% % \footnote{We tackle constrains with constrained optimization (Appendix \ref{sec:constrained_opt_appendix}) and do not explicitly show the constrain in the following.}
% ensures that the prediction is unchanged.
% and $\bx$ is similar to its perturbed version $\bx+\delta$. 
% \textcolor{red}{Section 4 deals with constraints.}
The constraint can make manipulating feature ranking more difficult. 

\section{Explanation Robustness via Thickness}
\label{sec:thickness_whole_section}
% \textcolor{red}{Do we have the additional advantage of being on the Pareto front?}
\subsection{Ranking explanation thickness and surrogates}
\label{sec:ranking_thickness_def}

\noindent \textbf{Quantify the gap.}
Given a model $f$ and the associated original explanation $\mathcal{I}(\mathbf{x})$ with respect to an instance $\mathbf{x} \in \mathbb{R}^n$,
% we denote $\mathcal{I}(\mathbf{x})_i$ as the feature importance of the $i$-th feature,
% which attributes the model output by Taylor first-order expansion.
% A feature with negative importance, 
% $\mathcal{I}(\mathbf{x})_i<0$,
% has a negative impact to the prediction of the class $c$.
we denote the \textit{gap} between the importance of $i$-th and $j$-th features by
$h(\mathbf{x}, i, j) = \mathcal{I} (\mathbf{x})_i - \mathcal{I} (\mathbf{x})_j$. 
% Again, 
% $\mathcal{R}(\mathbf{x})_i$ indicates the index of the $i$-th most important feature in $\mathcal{I}(\mathbf{x})$.
Clearly, 
$h(\mathbf{x},i,j)>0$ \textit{if and only if}
% when the $i$-th feature has a greater gradient magnitude, 
% thus more important
% than the $j$-th feature, 
when the $i$-th feature has a more significantly positive contribution to the prediction than the $j$-th feature.
Although the feature importance order varies across different $\mathbf{x}$,
for notation simplicity,
% for brevity,
we label the features in descending order such that the $i$-th feature is always more important than the $j$-th one,
namely,
$h(\mathbf{x},i,j)>0$,
$\forall i < j$,
with respect to the \textit{original} input $\mathbf{x}$.
% we assume that all original inputs have the same feature rankings for simplicity.
This assumption will not affect the following analysis.
% \textcolor{red}{Some motivation is needed before this definition.}

\textbf{Ranking robustness.}
Ranking robustness \cite{goren2018ranking,zhou2006ranking}
measures how much the ranking changes concerning slight input perturbations.
Now, we consider the perturbed input $\bx^\prime =\bx+\delta$ 
% where $\bx^\prime$ can be drawn from arbitrary distributions and $\|\delta\|_2\leq \epsilon$.
with $\|\delta\|_2\leq \epsilon$.
Apparently,
the adversary in Eq. (\ref{eq:attack_general}) tries to flip the ranking between $(i,j)$ features such that $h(\bx^\prime,i,j)<0$ for some $i<j$.
Meanwhile,
a model with robust ranking explanation is supposed to retain the rankings between any two features, $h(\bx^\prime, i,j)>0 ,\forall i<j$.
We define the \textit{explanation} thickness by the probability of the relative ranking of the features ($i,j$) being unchanged in a neighborhood of $\bx$.
% 
% As mentioned in Sec. \ref{sec:preliminary},
% Ranking Robustness is evaluated on both $\mathcal{I}(\bx)$ and $\mathcal{I}(\bx^\prime)$,
% and the gap for \textit{original} input $\bx$ is insufficient.
% % Thus, we consider an expectation over the possible perturbed input $\bx^\prime =\bx+\delta$.
% % 
The relevant work \cite{yang2020boundary} proposes the boundary thickness to evaluate a model's \textit{prediction} robustness by measuring the expected distance between two level sets.
% boundary thickness measures the expected distance between two \textit{level sets} from \textit{different classes} by multiplying the probability of samples within the two level sets and the distance between two level sets.
% Formally,
% boundary thickness is 
% defined as $\mathbbm{E}_{\bx^i,\bx^j}\left[ \|\bx^i - \bx^j \| \int_0^1 \mathds{1} [ \alpha < g_{ij}(\bx(t)) < \beta ] dt \right]$,
% where $\alpha$ and $\beta$ specify two level sets of interest.
% Two samples, $\bx^i$ and $\bx^j$, are predicted with class $i$ and $j$, respectively.
% $g_{ij}(\bx)=f(\bx)_i-f(\bx)_j$ and $\bx(t)= t\bx^i + (1-t) \bx^j$.
% Instead,
% we propose the ranking thickness to measure the \textit{explanation} robustness.
% It is the probability measure of the neighborhood of $\bx$ where the relative ranking of the features ($i,j$) is unchanged.
% we measure the \textit{explanation} robustness by the probability of the relative ranking of the features ($i,j$) being unchanged in a neighborhood of $\bx$.

\begin{definition}[Local Pairwise Ranking Thickness]
% For a distribution $\mathcal{D}$ over pairs of samples $(\mathbf{x}_r, \mathbf{x}_s)\sim \mathcal{D}$,
% the ranking thickness of a prediction function $f$ with respect to the $i$-th and $j$-th features
% is
Given a model $f$,
an input $\mathbf{x} \in \mathcal{X}$ 
% drawn from sample space 
and a distribution $\mathcal{D}$ over $\mathbf{x}^\prime$,
the local pairwise ranking thickness (as a probability)
% the $i$-th and $j$-th 
of the pair of features ($i,j$) is
\begin{equation}
\label{eq:pairwise_rank_thick}
\small
    % \Theta(f,\mathcal{D},i,j) 
    % = \mathbb{E}_{(\mathbf{x}_r,\mathbf{x}_s)\sim \mathcal{D}}
    \tilde{\Theta}(f, \mathbf{x}, \mathcal{D}, i, j)
    \defeq \mathbb{E}_{\mathbf{x}^\prime \sim \mathcal{D}}
    \left[ \int_0^1 
    \mathds{1} [h(\mathbf{x}(t), i,j) \geq 0] dt \right],
\end{equation}
% \textcolor{red}{For the indicator function, use $\mathds{1}[\cdot]$.}
where 
$\mathbf{x}^\prime \sim \mathcal{D}$ 
is the perturbed input drawn from a neighborhood of $\mathbf{x}$.
$\mathbf{x}(t)=(1-t)\mathbf{x} + t\mathbf{x}^\prime, t\in [0,1],$ is on the line segment connecting the sample pair ($\mathbf{x}, \mathbf{x}^\prime$).
% \textcolor{red}{$x_s$ should be defined here too.}
\end{definition}

Clearly,
$\tilde{\Theta} \leq 1$ and the equality holds when the relative importance of the $i$-th and $j$-th features is never flipped.
% The integration calculates the length of the line segment where the $i$-th feature is more important than $j$-th feature.
The integration calculates the probability that the $i$-th feature is more important than $j$-th feature
between $\mathbf{x}$ and $\mathbf{x}^\prime$.
The expectation considers all such probabilities
% line segments 
where
$\mathbf{x}^\prime$ is drawn from a Uniform distribution $U(\mathbf{x},\epsilon)$  \cite{wang2020smoothed}
or a Gaussian distribution $\mathcal{N}(\mathbf{x}, \sigma^2 I)$ 
\cite{smilkov2017smoothgrad}.
% \cite{smilkov2017smoothgrad,wang2020smoothed}.
The expectation makes the thickness estimation more comprehensive around the neighborhood of $\bx$. 
% resulting in a stable measurement of the thickness.
Alternatively,
we can set $\mathbf{x}^\prime$ to an adversarial sample local to $\mathbf{x}$ \cite{yang2020boundary} to find the worst case of the thickness,
and we will further discuss it in Sec. \ref{sec:ranking_thickness_defense}.

\noindent \textbf{Relaxation.}
Due to the non-differentiability of the indicator function in Eq. (\ref{eq:pairwise_rank_thick}),
it is difficult to analyze and optimize the thickness efficiently. 
Alternatively,
we remove the indicator function 
% of $h$, such as $h(\mathbf{x},i,j)$ itself.
% or $\exp\{h(\mathbf{x},i,j)\}$.
and define the local pairwise ranking thickness as
\begin{equation}
\label{eq:relax_pairwise_thickness}
    \Theta(f, \mathbf{x}, \mathcal{D}, i, j)
    \defeq \mathbb{E}_{\mathbf{x}^\prime \sim \mathcal{D}}
    \left[ \int_0^1 
    h(\mathbf{x}(t), i,j) dt \right],
\end{equation}
which is still monotonically increasing in $h$.
$\Theta(f,\bx,\mathcal{D},i,j)$ measures the \textit{expected gap} between the importance score of $i$ and $j$ features.
% It is highly positively related to $\tilde{\Theta}$,
% as a larger gap between features prevent the flip of rankings.
Our following analysis will focus on $\Theta$.

\noindent \textbf{Top-$k$ thickness.}
Existing works in general robust ranking propose maintaining the ranking 
% of \textit{every} pair 
between \textit{every} two features \cite{zhou2021practical},
demanding a complexity of $\mathcal{O}(n^2)$ for $n$ features.
However, as shown in Fig. \ref{fig:intro_example},
only the top-$k$ important features in $\mathcal{I}(\mathbf{x})$ and the robustness of their positions are more relevant to human perception of explanation.
Thus, only the relative ranking
% \textit{between} any two features from the top-$k$ and the remaining ones are relevant.
between a feature from the top-$k$ and another one from the remaining ones are relevant.
% Consequently,
% as Eq. (\ref{eq:topk_rank_thick}) shows,
We define the following robustness metric that requires nearly-linear complexity $\mathcal{O}(n)$ when $k \ll n$.

\begin{definition}[Local Top-$k$ Ranking Thickness]
% For a distribution $\mathcal{D}$ over pairs of samples $(\mathbf{x}_r, \mathbf{x}_s)\sim \mathcal{D}$,
% the ranking thickness of a prediction function $f$ with respect to the top-$k$ features in a given explanation 
% is
Given a model $f$, an input $\mathbf{x} \in \mathcal{X}$, 
% drawn from sample space,
and a distribution $\mathcal{D}$ over $\mathbf{x}^\prime$,
the local thickness of the ranking of the top-$k$ features is
\begin{equation}
\label{eq:topk_rank_thick}
\begin{aligned}
    \Theta(f,\mathbf{x},\mathcal{D};k, n)
    & \defeq
    % \sum_{i\in\{1,\dots,k\}}\sum_{j\in\{k+1,\dots,n\}}
    % \mathbb{E}_{\mathbf{x}_r} \left[
    \frac{1}{m}
    \sum_{i=1}^k \sum_{j=k+1}^n
    \Theta(f, \mathbf{x}, \mathcal{D},i,j),
    %\\ & 
    % \right]
    % =
    % \mathbb{E}_{
    % % \mathbf{x}_r,
    % \mathbf{x}_s\sim \mathcal{D}}
    % \left[
    % % \sum_{i\in\{1,\dots,k\}}\sum_{j\in\{k+1,\dots,n\}}
    % \int_0^1 
    % \frac{1}{N}
    % \sum_{i=1}^k \sum_{j=k+1}^n
    % \mathds{1} [h(\mathbf{x}(t), i,j) \geq 0] dt \right],
\end{aligned}
\end{equation}
where 
% $n$ is the number of features and 
$m=k(n-k)$,
% is for normalization,
and $\tilde{\Theta}(f,\bx,\mathcal{D};k,n)$ with an indicator function can be defined in a similar way.
% such that $\Theta(f,\mathbf{x}, \mathcal{D}; k, n)$ 
% which is within the range of $[0,1]$.
\end{definition}

\subsection{Training for robust ranking explanations}
\label{sec:ranking_thickness_defense}
% The theory in Sec.~\ref{sec:analysis} shows that maximizing the ranking explanation thickness can make attacks more difficult and thus the explanation more robust.
To make attacks more difficult and thus the explanation more robust,
we add $\Theta(f,\mathbf{x}, \mathcal{D};k,n)$ as a regularizer when training $f$ on the training set $(\mathcal{X}_T,\mathcal{Y}_T)$:
\begin{equation*}
    \min_{\bw} \mathcal{L}_{total}(f) =
    % \mathcal{L}_{cls} (f;\mathcal{X}_T, \mathcal{Y}_T) 
    \mathcal{L}_{cls} (f) 
    - \lambda
    \mathbb{E}_{\mathbf{x}\in\mathcal{X}_T}
    \left[
    \Theta(f,\mathbf{x}, \mathcal{D};k,n)
    \right],
\end{equation*}
where
$\mathcal{L}_{cls}(f)$ is the empirical classification loss of $f(\bx, \bw)$ and $\lambda>0$ is a hyperparameter.

\noindent \textbf{A surrogate bound of explanation thickness.}
% \textcolor{red}{The following paragraph discusses the challenges of solving the above optimization.}
% Similar to dealing with the non-differentiable dilemma in attacks, 
% we replace the indicator function with a monotonically increasing function and drop the normalization term $N$.
% Directly optimizing $\Theta$ is difficult due to the expectation and integration
% that defines $\Theta(f,\mathbf{x}, \mathcal{D},k;n)$ in Eq. (\ref{eq:topk_rank_thick}).
% Sampling the perturbed input $\bx^\prime$ from the uniform or Gaussian distribution around $\mathbf{x}$ is expensive and has a large variance when $\mathcal{X}$ is high-dimensional~\cite{smilkov2017smoothgrad,wang2020smoothed}.
% Evaluating the gradient of $h(\mathbf{x},i,j)=\mathcal{I}(\mathbf{x})_i-\mathcal{I}(\mathbf{x})_j$ requires second-order derivatives and at least two backward propagation per evaluation~\cite{pearlmutter1994fast},
% and the cost is multiplied by the number of samples $\mathbf{x}(t)$ on the line segment between $\bx$ and $\bx^\prime$ for any given $\bx^\prime$.
% % Training models with such sampling schema has an unacceptable time complexity,
% % since it requires calling backward propagation tremendous times for many neighboring samples $\mathbf{x}_s$ concerning every $\mathbf{x}_r$.
% We instead derive and maximize a lower bound of $\Theta(f,\mathbf{x}, \mathcal{D},k;n)$ avoid sampling $\mathbf{x}^\prime$ and $\bx(t)$.
% % in terms of explanation ranking.
Directly optimizing $\Theta$ in Eq. (\ref{eq:topk_rank_thick})
requires $M_1 \times M_2 \times 2$ backward propagations \textit{per} training sample \textit{per} iteration.
$M_1$ is the number of perturbed samples
% $\bx^\prime \sim \mathcal{D}$,
sampled from $U(\bx,\epsilon)$ or $\mathcal{N}(\bx, \sigma^2I)$, 
or the number of iterations for finding the adversarial sample $\bx^\prime$;
$M_2$ is the number of interpolations $\mathbf{x}(t)$ sampled from the line segment between $\bx$ and $\bx^\prime$;
and evaluating the gradient of $h(\mathbf{x},i,j)=\mathcal{I}(\mathbf{x})_i-\mathcal{I}(\mathbf{x})_j$ requires
% second-order derivatives and 
at least $2$ backward propagations~\cite{pearlmutter1994fast}.
To avoid sampling $\bx^\prime$ and $\bx(t)$,
we derive a lower bound of $\Theta$, which
requires only $2$ backward propagations to maximize. 
% \textit{per} $\bx$ \textit{per} iteration. 

\begin{definition}[Locally Lipschitz continuity]
\label{definition:locally_lipschitz}
A function $f$ is $L$-locally Lipschitz continuous if 
% for every $\mathbf{x}$,
$\|f(\mathbf{x})-f(\mathbf{x}^\prime)\|_2 \leq L \|\mathbf{x}-\mathbf{x}^\prime \|_2$
holds for all $\mathbf{x}^\prime \in \mathcal{B}_2(\mathbf{x},\epsilon) $
$=\{ \mathbf{x}^\prime \in \mathbb{R}^n: \|\mathbf{x}-\mathbf{x}^\prime\|_2 \leq \epsilon \}$.
\end{definition}

% \textcolor{red}{Before any formal statement, such as a proposition, theorem, etc. , you need to use plain English to describe the motivation/background/contexts of the statement. 
% For example, it is hard to understand why Proposition 1 (the bounds) is useful. 
% After the formal statement, add comments to the statement to deepen readers' understanding of the statement. For example, what the bounds in Proposition 1 tell us about the properties of the functions and robustness.}

\begin{proposition}{\textnormal{(Bounds of thickness)}}
\label{prop:bounds_of_thickness}
Given a $L$-locally Lipschitz model $f$,
% $f(\mathbf{x})$,
for some $L$,
% (
% $L \geq \frac{ \|\mathbf{x}-\mathbf{x}^\ast\|_2 * \max_i \max_{\mathbf{x}^\prime\in B_2(\mathbf{x},\epsilon)} 
% \|H(\mathbf{x}^\prime)_i\|_2}{2}$ and $\mathbf{x}^\ast = \argmax_{\mathbf{x}^\prime \in B_2(\mathbf{x},\epsilon)}\frac{\|\mathcal{I}(\mathbf{x}) - \mathcal{I}(\mathbf{x}^\prime)\|_2}{\|\mathbf{x}-\mathbf{x}^\prime\|}$,
% \textit{[Plan to put it in the appendix]}
% )
% then
% \textit{relaxed} 
local pairwise ranking thickness
$\Theta(f, \mathbf{x}, \mathcal{D}, i, j)$
% for the ($i,j$) pair of target $\mathbf{x}$ 
% = 
% \mathbb{E}_{\mathbf{x}^\prime \sim \mathcal{D}} \left[ \int_0^1 
% h(\mathbf{x}(t), i,j) \geq 0 dt \right]$
is bounded by 
\begin{equation}
\begin{aligned}
    & h(\mathbf{x},i,j) - 
    \epsilon * \frac{1}{2} \|H(\mathbf{x})_i -H(\mathbf{x})_j\|_2
    \leq \\
    % & \mathbb{E}_{\mathbf{x}^\prime} 
    % \left[ \int_0^1 h(\mathbf{x}(t),i,j) dt
    % \right] 
    &
    \qquad
    \Theta(f, \mathbf{x}, \mathcal{D}, i, j)
    \leq
    h(\mathbf{x},i,j) +
    \epsilon (L_i + L_j),
    \label{eq:upper_lower_thickness}
\end{aligned}
\end{equation}
where $H(\mathbf{x})_i$ is the derivative of $\mathcal{I}(\mathbf{x})_i$ with respect to the input $\bx$, 
% with respect to 
% the input 
% $\mathbf{x}$,
and 
$L_i=\max_{\mathbf{x}^\prime \in \mathcal{B}_2(\mathbf{x},\epsilon)} \| H(\mathbf{x}^\prime)_i \|_2$.
\end{proposition}
% \end{theorem}

% \begin{prop}
% [Option 2: starting from locally Lipschitz $\mathcal{I}(\mathbf{x})_i,\forall i$] 
% Given a model $f(\mathbf{x})$ whose gradient $\mathcal{I}(\mathbf{x})_i$ is $L_i$-locally Lipschitz
% for any $i$, 
% % where $L=\max_i L_i$,
% the \textit{relaxed} pairwise ranking thickness for ($i,j$) pair of a target $\mathbf{x}_r$ is bounded by 
% \begin{equation}
%     % g(\mathbf{x}_r)_i - g(\mathbf{x}_r)_j -
%     h(\mathbf{x}_r,i,j) - 
%     \epsilon * \frac{1}{2} \|H(\mathbf{x}_r)_i -H(\mathbf{x}_r)_j\|_2
%     \leq 
%     \mathbb{E}_{\mathbf{x}_s}\left[
%     \int_0^1 h(\mathbf{x}(t),i,j) dt
%     \right]
%     \leq
%     % g(\mathbf{x}_r)_i - g(\mathbf{x}_r)_j +
%     h(\mathbf{x}_r,i,j) +
%     \epsilon * (L_i + L_j),
% \end{equation}
% where $H(\mathbf{x})_i$ as the $i$-th column of Hessian matrix of $f$ with respect to the input $\mathbf{x}$,
% and 
% $L_i=\max_{\mathbf{x}^\prime \in B_2(\mathbf{x},\epsilon)} \| H(\mathbf{x}^\prime)_i \|_2$.
% \end{prop}
Note that the bounds are related to 
% the gap between the importance of the ($i,j$) pair of features, 
$h(\mathbf{x},i,j)=\mathcal{I}(\bx)_i-\mathcal{I}(\bx)_j$
and the Hessian of $f$.
% The first term, 
% $h(\mathbf{x},i,j)=\mathcal{I}(\bx)_i-\mathcal{I}(\bx)_j$,
% is the gap between the importance of the ($i,j$) pair of features,
% and the second terms on both sides are related to the Hessian matrix.
The bounds of the local top-$k$ ranking thickness could be derived similarly. 
% The detailed proofs and the similar bounds of the local top-$k$ ranking thickness are in Appendix \ref{sec:appendix-proof-bounds}. 

The bounds have the following implications.
\begin{itemize}[leftmargin=.1in]
    \item
    The bounds are 
    related to $\mathbf{x}$, not $\mathbf{x}^\prime$,
    and optimizing the bounds requires $M_1\times M_2$ \textit{fewer} times of backward propagations and frees from heavy computations
    in Eq. (\ref{eq:relax_pairwise_thickness}).
    \item 
The bounds are related to the perturbation budget $\epsilon$, but not to the distribution $\mathcal{D}$.
    Thus,
    Eq. (\ref{eq:upper_lower_thickness}) is valid for adversarial sample $\mathbf{x}^\prime$ \cite{yang2020boundary} with $\|\delta\|_2\leq \epsilon$ or any random distribution 
    % within the norm ball $\mathcal{B}_2(\mathbf{x},\epsilon)$.
    such as $U(\mathbf{x},\epsilon)$ \cite{wang2020smoothed}.
    \item 
    We reveal another motivation for minimizing Hessian norm:
    rather than smoothing the curvature, we aim to tighten the bounds of thickness and to ultimately maximize the thickness without dealing with neighbor sampling and line integration in Eq. (\ref{eq:relax_pairwise_thickness}). 
    In particular,
    as $\| H(\mathbf{x}) \|_2 \to 0$,
    we have $L_i+L_j \to 0$, 
    $\|H_i(\mathbf{x})-H_j(\mathbf{x})\|_2 \to 0$, and
    % in both sides,
    $\lim_{\|H(\mathbf{x}) \|_2 \to 0} \mathbb{E}_{\mathbf{x}^\prime} \left[ \int_0^1 h(\mathbf{x}(t),i,j)dt \right] = h(\mathbf{x},i,j)$.
    \item 
    The bounds are related to $h(\mathbf{x},i,j)=\mathcal{I}_i(\bx)-\mathcal{I}_j(\bx)$ and the Hessian of $f$.
    Thus,
    \textit{only} minimizing a Hessian norm~\cite{dombrowski2019explanations,dombrowski2021towards} is insufficient for ranking explanation robustness (see Table \ref{tab:comprehensive_result}). 
    \item 
    The most relevant prior work is~\cite{hein2017formal}. 
    They connect the robustness of prediction $f_c$ to the ratio $\max_j \frac{f(\mathbf{x})_c-f(\mathbf{x})_j}{
    % \max_{\mathbf{x}^\prime \in B_2(\mathbf{x},\epsilon)} 
    \| \mathcal{I}(\mathbf{x}^\prime, c) - \mathcal{I}(\mathbf{x}^\prime, j) \|_2 }$, 
    % Similar to~\cite{moosavi2019robustness}, 
    where the ratio stems from the \textit{optimal} perturbation direction.
    Although we can adopt their proof to obtain a similar ratio $\frac{\mathcal{I}(\mathbf{x})_i-\mathcal{I}(\mathbf{x})_j}
    {\| \nabla \mathcal{I}(\mathbf{x}^\prime)_i - \nabla \mathcal{I}(\mathbf{x}^\prime)_j \|_2 }$, 
    our experiments show that the second-order term in the denominator makes the optimization less stable and the training can hardly converge 
    % (see Appendix \ref{sec:sensitivity_analysis_experiments}).
    (see Table \ref{tab:sensitivity_analysis_pretrain}).
\end{itemize}

% which indicates the local curvature around $\mathbf{x}_r$. 
% The result is consistent with intuition that the thickness is bounded by the gradient gap at point $\mathbf{x}_r$ as well as the curvature, 
% depicted by Hessian matrix, 
% around the point $\mathbf{x}_r$.
% \textcolor{red}{Another thing to mention is the connection between the bounds and the Hessian-based regularization in our robust training algorithm.}
% Inspired by Proposition 1,
% we know that $h(\mathbf{x}(t),i,j)$ is bounded by the gap on $\mathbf{x}_r$ and a Hessian-related terms nearby $\mathbf{x}_r$ given a $L$-locally Lipschitz function $f$. 
% In an extreme case that $H=\mathbf{0}$,
% the Hessian-related terms in both size of Eq. (\ref{eq:upper_lower_thickness}) equal to zeros, 
% and based on the Squeeze theorem we have $\mathbb{E}_{\mathbf{x}_s} \left[ \int_0^1 h(\mathbf{x}(t),i,j)dt \right] 
% = h(\mathbf{x}_r,i,j)$.
Based on Prop. \ref{prop:bounds_of_thickness}, 
simultaneously maximizing the gap and minimizing Hessian norm improve the thickness.
Thus, 
we have the following optimization problem for training an accurate classifier with robust feature ranking
% salient features as 
explanations:
\begin{equation}
\begin{aligned}
\label{eq:defense_thick}
    \min_\bw \mathcal{L}_{total}(f) =
    % \mathcal{L}_{cls}(f;\mathcal{X}_T,\mathcal{Y}_T)
    &
    \mathcal{L}_{cls}(f)
    - 
    \lambda_1
    \mathbb{E}_{\mathbf{x}}
    % \in \mathcal{X}_T}
    \left[
    \sum_{i=1}^k \sum_{j=k+1}^n
    h(\mathbf{x}, i,j) \right]   \\
    & +  \lambda_2
    \mathbb{E}_{\mathbf{x}}
    % \in \mathcal{X}_T} 
    \left[ 
    \| H(\mathbf{x}) \|_2 
    \right]
    ,
\end{aligned}
\end{equation}
where 
% $f=f(\mathbf{x};\bw)$ and 
$\lambda_1, \lambda_2 \geq 0$.
% A significant merit is that 
In this way,
we optimize the gap and Hessian norm concentrating in $\mathbf{x}\in \mathcal{X}_T$ and are free from expensive sampling. 
% \cite{wang2020smoothed}.
When $\lambda_1=0$,
we recover Hessian norm minimization to smooth the curvature~\cite{dombrowski2021towards}.
When $\lambda_2=0$, we only increase the gaps.
We call the strategy that trains $f$ using Eq. (\ref{eq:defense_thick}) \textit{Robust Ranking Explanation via Thickness} (\textbf{R2ET}).
% Authors in \cite{hein2017formal} propose a similar regularization term for $\lambda_2=0$,
% however,
% for a distinct goal.
% They identify that model predictions do not change under any attacks with no more than
% $\max_j \frac{f(\mathbf{x})_c-f(\mathbf{x})_j}{
% % \max_{\mathbf{x}^\prime \in B_2(\mathbf{x},\epsilon)} 
% \| \mathcal{I}(\mathbf{x}^\prime)_c - \mathcal{I}(\mathbf{x}^\prime)_j \|_2 }$ perturbations.
% Considering that $\mathcal{L}_{cls}$ already maximize the numerator,
% they \textit{minimize} the proxy denominator $\| \mathcal{I}(\mathbf{x})_i - \mathcal{I}(\mathbf{x})_j \|_2^2$ for any $(i,j)$ pair.
% On the contrary,
% improving the explanation ranking thickness requires the \textit{maximization} of $h(\mathbf{x}_r,i,j)$.
% One may consider $\frac{\mathcal{I}(\mathbf{x})_c-\mathcal{I}(\mathbf{x})_j}{
% \| \nabla \mathcal{I}(\mathbf{x}^\prime)_c - \nabla \mathcal{I}(\mathbf{x}^\prime)_j \|_2 }$ as the regularizer to boost the explanation ranking thickness.
% However, 
% the computation of the denominator is costly,
% and we find it is hard to smoothly optimize the Hessian-related terms in the denominator empirically.

% \textcolor{red}{Add a paragraph to discuss how the above challenges are resolved, including how Propositions 1 and 2 are used to formulate the final optimization problem with both $h$ and the Hessian norm minimization.
% How the Hessian norm related to the norm of the row difference of the Hessian are related?}

\noindent \textbf{Connection to adversarial training (AT).}
We have following proposition based on the prior work~\cite{xu2009robustness}:
\begin{proposition}
\label{prop:r2et_at}
The optimization problem in Eq. (\ref{eq:defense_thick}) is equivalent to the following min-max problem:
\begin{equation}
\small
\begin{aligned}
\label{eq:obj_at}
    \min_\bw \max_{(\delta_{1,k+1},\dots,\delta_{k,n}) \in \mathcal{N}} 
    \mathcal{L}_{cls}
    - 
    \mathbb{E}_\bx \left[
    \sum_{i=1}^k\sum_{j=k+1}^n h(\bx + \delta_{i,j}, i, j)
    \right],
\end{aligned}
\end{equation}
where $\delta_{i,j}$ is a perturbation to $\bx$ targeting the $(i,j)$ pair of features.
$\mathcal{N}$ is the feasible set of perturbations where \textit{each} $\delta_{i,j}$ is independent of each other, with
% but the aggregation of perturbations are controlled,
% e.g., 
% the norm of the summation of perturbations is not larger than $\epsilon$
$\|\sum_{i,j} \delta_{i,j} \|\leq \epsilon$
\end{proposition}
% The goal of AT, in our case, is to train a model with robust ranking \textit{explanations} by a min-max game:
% $\min_\bw \max_{(\delta_{1,k+1},\dots,\delta_{k,n}) \in \mathcal{N}} \mathcal{L}_{cls} - \sum_{i=0}^k\sum_{j=k+1}^n h(\bx + \delta_{i,j}, i, j)$.
% we prove the \textit{equivalence} between the objectives of R2ET in Eq. (\ref{eq:defense_thick}) and AT.
% In other words,
% See Appendix \ref{sec:appendix_connect_at_thickness} for the formal definitions and proof inspired by 
% % the previous work
% \cite{xu2009robustness}.
Prop. \ref{prop:r2et_at} indicates that R2ET has the same effect as the AT but uses a regularization to bypass heavy computations.
Specifically, 
% as discussed earlier,
the bottleneck of the above AT is the high time complexity for finding the optimal $\delta$ for any $\bx$,
% and training $f$,
which makes the AT $M_1$ times more expensive than R2ET,
where $M_1$ is the number of attack iterations per $\bx$.
% Consequently, R2ET is scalable to large-scale datasets than AT

% Adversarial Training (AT) finds the optimal $\delta^\ast$ that manipulates the feature ranking by minimizing 
% % the objective 
% % $\mathbb{E}_{
% %     % \mathbf{x}_r,
% %     \mathbf{x}^\prime=\mathbf{x}+\delta^\ast}
% % \left[
% $\sum_{i=1}^k \sum_{j=k+1}^n
% h(\mathbf{x}+\delta^\ast, i,j)$.
% % \right]$.
% The objective is not greater than the lower bounds of the thickness
% $\mathbb{E}_{\mathbf{x}^\prime \sim \mathcal{D}} 
% \left[ \int_0^1 \sum_{i=1}^k \sum_{j=k+1}^n h(\mathbf{x}(t), i,j) dt \right]$
% and thus AT can also optimize thickness.
% However, 
% the bottleneck is the high time complexity for finding $\delta^\ast$ for any $\bx$ and training $f$.
% Specifically,
% substituting $\bx^\prime$ with $\bx+\delta^\ast$ requires $M_1 * M_2$ higher time complexity than R2ET for $M_1$ attack iterations and $M_2$ interpolation samples.
% Alternatively,
% replacing $\bx(t)$ with $\bx+\delta^\ast$ demands $M_1$ higher time complexity but it is a coarser estimation of the thickness.
% % and we cannot finish AT on some of the larger datasets.

\noindent \textbf{Selecting pairs to compare.}
% We can speed up the calculation of R2ET by comparing fewer feature pairs.
% In Eq. (\ref{eq:defense_thick}),
% we enumerate $k*(n-k)$ pairs between top-$k$ and bottom $(n-k)$ features.
Notice that the features in the long tail are less likely to be confused with the top-$k$ features, 
% (visualized in Appendix \ref{sec:sensitivity_analysis_experiments}),
% (visualized in Fig. \ref{fig:gradient_magnitude}),
we set the $k$-th salient feature as the ``anchor'', 
% and only maintain the relative ranking between the anchor and other top-$2k$ features. 
and approximate the top-$k$ ranking thickness by $k^\prime$ pairs of features $\sum_{i=1}^{k^\prime} h(\mathbf{x},k-i,k+i)$ with $\mathcal{O}(k^\prime)$ complexity.
When $k^\prime=k$, we preserve the relative rankings of the top-$2k$ features,
which is named \textbf{R2ET-mm}.

\section{Experiments}
\label{sec:experiemnt_whole_section}
\subsection{Experimental Settings}
\noindent \textbf{Model architectures and datasets.}
We adopt two types of network architectures: single-input DNNs and dual-input Siamese Networks (SNs).
For single-input DNN,
we use three tabular datasets: Bank \cite{moro2014data},
Adult, and COMPAS \cite{mothilal2020explaining}, 
and an image dataset CIFAR-10 \cite{krizhevsky2009learning} with ResNet.
For SNs that compare two inputs, 
we use the image dataset MNIST \cite{lecun1998gradient} and two graph datasets of brain networks: BP \cite{ma2019deep} and ADHD \cite{ma2016multi}.
In BP and ADHD, each brain network comprises 82 and 116 nodes, respectively. Since the datasets are limited, we employ five-fold cross-validation, where three folds are used for training, one for validation, and one for testing. We create training sets by pairing any two training graphs. 
We pair any two training graphs as the training set.
To simulate real medical diagnosis (by comparing a new sample with those in the database), 
each pair consists of a training graph and a validation (or test) graph as validation (or testing) sets.

% Additionally, each validation graph is paired with two training graphs,
% one from the same class and one from a different class.
% The test set adopts the same setting.

\noindent \textbf{Evaluation metrics.}
We use Precision@$k$ (P@$k$) \cite{ghorbani2019interpretation,wang2020smoothed}
% Top-$ k $ intersection ratio (P@$k$)
% Pearson correlation coefﬁcient (PCC) \cite{dombrowski2019explanations},
% and mean squared error (MSE) \cite{dombrowski2019explanations},
to quantify the similarity between two explanations before and after attacks.
% In addition to robustness, 
% In Appendix \ref{sec:constrained_opt_appendix} and \ref{sec:acc_robust_trade_off},
To ensure all the trained models have similarly good \textit{prediction} performance,
we guarantee that almost all the models have relatively high clean AUC.
We further keep the adversarial AUC high and sensitivity \cite{xu2020adversarial} close to zero when conducting attacks.
In Sec. \ref{sec:experiment_faithfulness},
% Concerning explanations \textit{faithfulness},
we report DFFOT \cite{serrano2019attention}, 
% measuring the minimum fraction of important features to be removed to flip the prediction,
comprehensiveness and sufficiency
\cite{deyoung2019eraser} to show that explanations from R2ET models are \textit{faithful}.
% measuring the changes of predictions before and after removing the most (or least, resp.) important features.
% The formal definitions and corresponding 
% Detailed results are disclosed in Appendix \ref{sec:append_experiments}.
\begin{itemize}[leftmargin=*]
    \item \textbf{Decision Flip - Fraction of Tokens (DFFOT)} \cite{serrano2019attention} measures the minimum fraction of important features to be removed to flip the prediction.
    % A \textit{lower} DFFOT indicates a more faithful explanation.
    Formally,
    % \begin{equation*}
    % \begin{aligned}
    % & \textnormal{DFFOT} = \min_k \frac{k}{n}, \\
    %     % \quad 
    %     & \textnormal{s.t.}
    %     \quad
    %     \argmax_c f(\bx)_c \neq \argmax_c f(\bx_{[\backslash k]})_c,
    % \end{aligned}
    % \end{equation*}
    % $ \textnormal{DFFOT} = 
    $\min_k \frac{k}{n},
        \textnormal{s.t.}
        \argmax_c f(\bx)_c \neq \argmax_c f(\bx_{[\backslash k]})_c,
    $
    where
    % $n$ is the number of features, and 
    $\bx_{[\backslash k]}$ is the perturbed input whose top-$k$ important features are removed.
    
    \item \textbf{Comprehensiveness (COMP)} \cite{deyoung2019eraser} 
    measures the changes in predictions before and after removing the most important features.
    % A \textit{higher} COMP means a more faithful explanation. 
    Formally,
    % \begin{equation*}
    %     \textnormal{COMP} = \frac{1}{\|K\|} \sum_{k\in K} |f(\bx)_c - f(\bx_{[\backslash k]})_c |,
    % \end{equation*}
    $
    % \textnormal{COMP} = 
    \frac{1}{\|K\|} \sum_{k\in K} |f(\bx)_c - f(\bx_{[\backslash k]})_c |$,
    where $K$ is $\{1,\dots,n\}$ for tabular data, 
    and  $\{1\% *n, 5\% *n, 10\% *n, 20\% *n, 50\% *n\}$ for images and graphs.
    
    \item \textbf{Sufficiency (SUFF)} \cite{deyoung2019eraser}
    measures the change of predictions if only the important tokens are preserved.
    % A \textit{lower} SUFF means a more faithful explanation.
    Formally,
    % \begin{equation*}
    %     \textnormal{SUFF} = \frac{1}{\|K\|} \sum_{k\in K} |f(\bx)_c - f(\bx_{[k]})_c |,
    % \end{equation*}
    $
    % \textnormal{SUFF} =
    \frac{1}{\|K\|} \sum_{k\in K} |f(\bx)_c - f(\bx_{[k]})_c |$,
    where $\bx_{[k]}$ is the perturbed input with only top-$k$ important features, 
    and $K$ is set the same as the one for COMP.
\end{itemize}

% \noindent \textbf{Running Environment.}
% We majorly conduct experiments for three tabular data on the following two machines.
% Both come with a 16-core Intel Xeon processor and four TITAN X GPUs.
% One machine installs 16.04.3 Ubuntu with 3.8.8 Python and 1.7.1 PyTorch,
% and the other installs 18.04.6 Ubuntu with 3.7.6 Python and 1.8.1 PyTorch.
% The image and graph datasets are run on a machine with two 10-core Intel Xeon processors and five GeForce RTX 2080 Ti GPUs, which installed 18.04.3 Ubuntu with 3.9.5 Python and 1.9.1 PyTorch.

\noindent \textbf{Explanation methods.}
% To evaluate the robustness of the explanations different training methods, 
We adopt SimpleGrad as the explanation method,
% if not specified. 
and similar conclusions to SimpleGrad can be drawn when adopting SmoothGrad \cite{smilkov2017smoothgrad} and Integrated Gradients  \cite{sundararajan2017axiomatic}.
% three gradient-based explanation methods, SimpleGrad, SmoothGrad (SG) \cite{smilkov2017smoothgrad}, and Integrated Gradients (IG) \cite{sundararajan2017axiomatic}.
% SimpleGrad simply uses the gradient of the models with respect to inputs as explanations as introduced in Sec. \ref{sec:preliminary}. 
% SG \cite{smilkov2017smoothgrad} takes random samples in a small neighborhood of input $\mathbf{x}$,
% and computes the average of the saliency maps for these random samples as the final results for $\mathbf{x}$. 
% IG \cite{sundararajan2017axiomatic} is defined as the integration of the saliency maps along a path from a reference point $\mathbf{x}^0$ to input $\mathbf{x}$.
% SimpleGrad will be used as explanation methods if not specified in this section. 
% The results for SG and IG are in Appendix \ref{sec:append_experiments}, and similar conclusions to SimpleGrad can be drawn.

\noindent \textbf{Hyperparamters.}
We pick $k=8$ for three tabular data,
$k=100$ for CIFAR-10,
and $k=50$ for MNIST and graphs.
We set the maximal training epoch as 300 for three tabular data,
100 for MNIST, 
and 10 for two graph datasets, BP and ADHD.
Almost all models are guaranteed to converge within given maximal epochs,
except when the regularization term weights are too large.
The learning rate is set to 1e-2 for three tabular data, 1e-3 for MNIST, 1e-2 for CIFAR-10, and 1e-4 for BP and ADHD.
% The early stop for three tabular data is 30, and no early stop is set for the image and graph datasets.
% Except in the sensitivity analysis in Appendix \ref{sec:sensitivity_analysis_experiments},
We set $k^\prime=k$ for R2ET and its variants for three tabular datasets and image datasets,
% Since for R2ET-mm and R2ET-$\textnormal{mm}_{\backslash H}$, 
% setting $k^\prime=k$ reduces cAUC on BP and ADHD, 
and $k^\prime=20$ for BP and ADHD.
% for both methods on both datasets.
% In each iteration,
% R2ET maintains the ranking from model in the previous iteration.
% 
As for attacks, 
we conduct attacks in a PGD-style \cite{madry2017towards} for at most 1000 iterations for tabular datasets, 
and perturb input with a 1e-3 budget in each iteration.
Thus, 
each input in tabular datasets can be manipulated with at most $\epsilon=10^{-3}*1000=1$ budget. 
The budget is set as $100 * 5e-2$ for CIFAR-10,
and $100 * 1e-2$ for the rest.
Inputs are normalized in image datasets such as CIFAR-10 and MNIST.

\subsection{Compared Methods}
\label{sec:method_to_compare}
We conduct two attacks in the PGD manner \cite{madry2017towards}:
Explanation Ranking attack (\textbf{ERAttack}) and \textbf{MSE attack}.
ERAttack minimizes $
    % \mathbb{E}_{\mathbf{x}_r}
    % \left[
    \sum_{i=1}^k \sum_{j=k+1}^n
    h(\mathbf{x}^\prime, i,j) 
    % \right]
    $
% in Eq. (\ref{eq:defense_thick}) 
to manipulate the ranking of features in explanation $\mathcal{I}(\bx)$,
and MSE attack maximizes the MSE (i.e., $\ell_2$ distance) between $\mathcal{I}(\mathbf{x})$ and $\mathcal{I}(\mathbf{x}^\prime)$.
% Both attacks are constrained to retain the target model's predictions by constrained optimization algorithms, see Appendix \ref{sec:append_experiments} for more details.
% We compare \textbf{ERAttack} with \textbf{MSE attack} whose goal is to maximize the MSE between $\mathcal{I}(\mathbf{x})$ and $\mathcal{I}(\mathbf{x}^\prime)$.
% We compare our defense strategies, 
% \textbf{R2ET} and its variants,
% with the state-of-the-art methods. 
We compare 
% Robust Ranking Explanation via Thickness
the proposed defense strategy \textbf{R2ET} 
% the proposed defense strategy optimizing Eq. (\ref{eq:defense_thick}),
with 
% the state-of-the-art methods. 
the following baselines.

\begin{itemize}[leftmargin=*]
  \setlength\itemsep{0.1em}
    \item \textbf{Vanilla}:
    provides the basic ReLU model trained without weight decay or any regularizer term. 
    \item \textbf{Weight decay (WD) \cite{dombrowski2021towards}}:
    uses weight decay during training to bound Hessian norm. 
    % and effectively improve the robustness of explanations.
    \item \textbf{Softplus (SP)} \cite{dombrowski2019explanations,dombrowski2021towards}:
    replaces ReLU with
    % Softplus as the activation function,
    % where 
    $\textnormal{Softplus}(x;\rho)=\frac{1}{\rho}\ln(1+e^{\rho x})$.
    % and a smaller $ \rho $ results in smoother curvature. 
    \item \textbf{Estimated-Hessian (Est-H)} \cite{dombrowski2021towards}:
    % adds the Hessian norm as part of the objective function,
    Hessian norm as the regularizer,
    which is estimated by the finite difference
    \cite{moosavi2019robustness}:
        $
        % \| H(\mathbf{x})\|_2 \approx
        \| \frac{\nabla f(\mathbf{x} + \kappa \mathbf{v})-\nabla f(\mathbf{x})}{\kappa}\|_2$,
    where $ \kappa \ll 1 $,
    $ \mathbf{v}=\frac{\textnormal{sign}(\nabla f(\mathbf{x}))}
    {||\textnormal{sign}(\nabla f(\mathbf{x}))||_2}$.
    It can be considered an ablation variant of R2ET ($\lambda_1=0$).
    \item \textbf{Exact-Hessian (Exact-H)}:
    the exact Hessian norm is used as the regularizer.
    % for datasets with few features.
    \item
    \textbf{SSR} \cite{wang2020smoothed}:
    sets the largest eigenvalue of the Hessian matrix as the regularizer.
    \item
    \textbf{Adversarial Training (AT) for robust prediction} 
    % \cite{huang2015learning,shaham2018understanding}: 
    \cite{huang2015learning,wong2020fast}: 
    % trains the model following a min-max
    % approach, e.g.,
    find $f$ by
    $\min_{f}
    \sum_{(\mathbf{x},y)\in(\mathcal{X}_T,\mathcal{Y}_T)}
    (\mathcal{L}_{cls} (f;\mathbf{x}+\delta^\ast, y)),$
    where
    $
    \delta^\ast = \argmax_\delta - \sum_{i=1}^k \sum_{j=k+1}^n
    h(\mathbf{x}+\delta, i,j).
    $
    \item
    \textbf{R2ET-mm}:
    selects \textit{multiple} distinct $i,j$ with \textit{minimal} $h(\mathbf{x},i,j)$ as discussed in Sec. \ref{sec:ranking_thickness_defense}.
    \item
    \textbf{R2ET}$_{\backslash \mathbf{H}}$ and 
    \textbf{R2ET-mm}$_{\backslash \mathbf{H}}$:
    They are the ablation variants of \textbf{R2ET} and \textbf{R2ET-mm}, respectively, without optimizing the Hessian-related term in Eq. (\ref{eq:defense_thick}) ($\lambda_2=0$).
\end{itemize}

\subsection{Overall robustness results}
\noindent \textbf{Attackability of ranking-based explanation.}
Table \ref{tab:comprehensive_result} reports P@$k$ under ERAttack and MSE attacks for every model on all datasets.
We observe that more than 50$\%$ of models achieve at least 90$\%$ P@$k$ under MSE attacks,
concluding that MSE attack cannot effectively alter the rankings of salient features,
even without extra defense (row Vanilla).
The ineffective attack method can give a false impression of explanation robustness, and a stronger attack is needed.  
ERAttack, on the other hand,
can remove more salient features from the top-$k$ positions for most models and datasets, 
leading to significantly lower P@$k$ values than MSE attack.
For example, 
% for the ADHD dataset,
% about $50*50\%=25$ salient features are out of top positions under the ERAttack on average.
about $50\%$ of the top $50$ features are out of top positions under ERAttack on average on ADHD,
while less than $20\%$ of the top $50$ features drop out of the top under the MSE attack.
% The trade-off between accuracy and explanation robustness is explored in 
% % Appendix \ref{sec:acc_robust_trade_off}.
% Fig. \ref{fig:acc_robust_trade_off}.
% Further results, 
% including that of ERAttack constrained retain
% the classifier's output,
% and the trade-off between accuracy and explanation robustness
% are in Appendix~\ref{sec:constrained_opt_appendix}
% and \ref{sec:acc_robust_trade_off}, respectively.

\noindent \textbf{Effectiveness of ranking thickness against ERAttacks.}
% Since there is a positive correlation between model's performance under MSE attack and ERAttack,
% \textcolor{red}{Somewhere in this paragraph, conclude that optimizing Hessian norm is not sufficient for robustness.}
We compare the performance of different defense strategies against ERAttack,
which is more effective than MSE attacks.
Similar conclusion can be made with MSE attacks case.
% (although similar patterns can be observed under the MSE attacks).
First, R2ET and its variants achieve the best (highest) top-$k$ explanation robustness for most datasets,
% (i.e., having the highest P@$k$ values), 
indicating R2ET methods' superiority for preserving the top salient features. 
% in the explanation under effective attacks.
Second,
it is counter-intuitive that 
$\textnormal{R2ET}_{\backslash H}$,
as an ablation version of R2ET,
outperforms R2ET on Adult and Bank.
The reason is that $\textnormal{R2ET}_{\backslash H}$ has a better ranking thickness on these datasets than R2ET
(see Fig. \ref{fig:correlation_manipulation_epoch} in Sec. \ref{sec:relation_thickness_other_metric}).
We conjecture that the number of features in the dataset can serve as a straightforward and intuitive indicator to determine the potential performance of R2ET or $\textnormal{R2ET}_{\backslash H}$ in practical scenarios.
Specifically,
in cases where the number of features is \textit{small}, such as in the Adult, Bank, and COMPAS datasets, 
it becomes easier to restrict the relative rankings among a limited set of features. 
As a result, both $\textnormal{R2ET}_{\backslash H}$ and $\textnormal{R2ET-mm}_{\backslash H}$ demonstrate good performance. 
In this case, reducing the Hessian norm diminishes the gap between features and adversely affects robustness, 
as observed when comparing Vanilla and Est-H or Exact-H.
Conversely, when the number of features is \textit{large}, it becomes significantly more challenging to maintain all the rankings solely by expanding the gaps between features, which is the approach taken by $\textnormal{R2ET}_{\backslash H}$ and $\textnormal{R2ET-mm}_{\backslash H}$. 
Due to the sheer number of features involved, R2ET, which simultaneously expands the gaps and minimizes the Hessian norm, has the potential to outperform the other methods.
The theoretical discussion is disclosed in Sec. \ref{sec:ranking_thickness_defense}.
% We will provide a more empirical analysis of this case in Sec. \ref{sec:relation_thickness_other_metric}.
Lastly, we consider the baselines that strive for a smoother curvature without considering the absolute gaps among feature importance,
including WD, SP, Est-H, Exact-H, and SSR.
% (\textcolor{red}{What are they?}).
Overall, their performance is unstable across datasets (SP on COMPAS and Est-H on BP).
However, the best performers always have the largest thickness 
% (\textcolor{red}{or gap?}) 
as demonstrated in Sec. \ref{sec:relation_thickness_other_metric}.
The above observations show that ranking thickness is a more fundamental measurement of ranking robustness.
% , confirming the theoretical analysis in Sec.~\ref{sec:analysis}. 
Besides that,
Est-H, Exact-H, and SSR
smooth the curvature by adding Hessian-related terms, 
either Hessian norm or the maximal eigenvalue of Hessian,
and they perform similarly.
Based on their performance in Table \ref{tab:comprehensive_result},
optimizing Hessian norm or its relevant terms solely is not sufficient to improve the ranking thickness.
% % \textcolor{red}{Did not explain two blank lines for Exact-H and SSR.}
% Exact-H uses the \textit{exact} Hessian norm as the regularization term, 
% which is infeasible in high-dimensional datasets.
% SSR \cite{wang2020smoothed} adopts the largest eigenvalues of the Hessian matrix as the regularization term,
% and the authors provide a closed-form formula to estimate the Hessian matrix for \textit{single-input} ReLU model.
% However, 
% we conduct experiments on \textit{dual-input} Siamese Networks for image and graph datasets, 
% and we cannot adopt a similar estimation.
Since it is extremely expensive to compute the exact Hessian norm and its eigenvalues,
both Exact-H and SSR are inapplicable to MNIST, CIFAR-10, ADHD and BP. 
We adopt ``fast''-AT \cite{wong2020fast} for AT baseline, where the inner maximization is solved by a single-step attack to balance the training time and robust performance.
``Fast''-AT, however, suffers from unstable robust performance as studied in \cite{li2020towards},
and cannot perform well in most datasets.

\begin{table*}[!tb]
\caption{\small P@$k$ (shown in percentage) of different robust models (rows) under \textbf{ERAttack} / \textbf{MSE attack}.
$k=8$ for the first three dataset, and $k=50$ for the rest.
Numbers in \textbf{bold} indicate the winner on the dataset, 
and \underline{numbers}  indicate the runner-up.
($\ast$ Est-H has 4.6\% and 3.9$\%$ lower clean AUC than R2ET-mm under two attacks, respectively, and is less useful in practice.)
}
\centering
\small
\begin{tabular}{ c || c c c | c c | c c}
\toprule
\textbf{Method} & Adult & Bank
& COMPAS & MNIST & CIFAR-10 & ADHD & BP \\
\midrule
Vanilla & 
87.6 / 87.7 & 83.0 / 94.0 & 84.2 / 99.7 &
59.0 / 64.0 & 66.5 / 68.3 &
45.5 / 81.1 & 69.4 / 88.9
\\
WD &
91.7 / 91.8 & 82.4 / 85.9 & 87.7 / 99.4 &
59.1 / 64.8 & 64.2 / 65.6 &
47.6 / 79.4 & 69.4 / 88.6
\\
SP & 
\underline{97.4} / \underline{97.5} & 95.4 / 95.5 & \textbf{99.5} / \textbf{100.0} &
62.9 / 66.9 & 67.2 / 71.9 &
42.5 / 81.3 & 68.7 / 90.1
\\
Est-H & 
87.1 / 87.2 & 78.4 / 81.8 & 82.6 / 97.7 &
85.2 / 90.2 & 77.1 / 78.7 &
58.2 / 83.7 & \textbf{75.0}$^\ast$ / \textbf{91.4}$^\ast$
\\
Exact-H & 
89.6 / 89.7 & 81.9 / 85.6 & 77.2 / 96.0 &
- / - & - / - & - / - & - / -
\\
SSR & 91.2 / 92.6 & 76.3 / 84.5 & 82.1 / 97.2 &
- / - & - / - & - / - & - / -
\\
AT & 68.4 / 91.4 & 80.0 / 88.4 &  84.2 / 90.5 
& 56.0 / 63.9 & 61.6 / 66.8 & 59.4 / 81.0 & 72.0 / 89.0
\\
\midrule
$\textnormal{R2ET}_{\backslash H}$ & 
\textbf{97.5} / \textbf{97.7} & \textbf{100.0} / \textbf{100.0} & 91.0 / 99.2 &
82.8 / 89.7 & 67.3 / 72.2 & 
60.7 / 86.8 & 70.9 / 89.5
\\
R2ET-$\textnormal{mm}_{\backslash H}$ &
93.5 / 93.6 & \underline{95.8} / \underline{98.2} & \underline{95.3} / 97.2 &
81.6 / 89.7 & \underline{77.7} / \textbf{79.4} &
\underline{64.2} / \underline{88.8} & 72.4 / 91.0
\\
% tx &
% \textbf{97.5} / \textbf{97.7} & \textbf{100.0} / \textbf{100.0} & 91.0 / 99.2 &
% 82.8 / 89.7 & 60.3 / 86.0 & 70.9 / 89.5
% \\
% etx & 
% 91.5 / 91.6 & 87.0 / 95.7 & 91.9 / 99.4 & 
% 75.2 / 86.1 & 60.7 / 86.2 & 72.1 / 89.5
% \\
% txmm &
% 93.5 / 93.6 & 95.8 / 98.2 & 95.3 / 97.2 &
% 81.6 / 89.7 & 64.2 / 87.5 & 73.1 / 90.7
% \\
% etxmm &
%  &  &  &
% 78.3 / 87.3 & 55.8 / 81.3 & 72.4 / 91.0
% \\
\midrule
R2ET & 
92.1 / 92.7 & 80.4 / 90.5 & 92.0 / \underline{99.9} &
\textbf{85.7} / \underline{90.8} & 75.0 / 77.4 & 
\textbf{71.6} / \textbf{91.3} & 71.5 / 89.9
\\
R2ET-mm &
87.8 / 87.9 & 75.1 / 85.4 & 82.1 / 98.4 &
\underline{85.3} / \textbf{91.4} & \textbf{78.0} / \underline{79.1} & 
58.8 / 87.5 & \underline{73.8} / \underline{91.1}
\\
% tx+Hess &
% 92.1 / 92.7 & 80.4 / 90.5 & 92.0 / 99.9 &
% 84.6 / 90.7 & \textbf{73.8} / 88.1 & 72.2 / 89.6
% \\
% etx+Hess &
%  &  &  &
% \textbf{85.7} / 90.8 & 72.0 / 86.6 & 71.5 / 89.9
% \\
% txmm+Hess &
% 87.8 / 87.9 & 75.1 / 85.4 & 82.1 / 98.4 &
% 84.4 / 89.3 & 59.3 / 90.6 & 73.8 / \textbf{91.7}
% \\
% etxmm+Hess & 
%  &  &  &
% 85.3 / \textbf{91.4} & 60.1 / \textbf{93.0} & 73.8 / 91.1
% \\
\bottomrule
\end{tabular}
\label{tab:comprehensive_result}
\end{table*}

% \subsection{A Closer Look at Ranking Thickness and Existing Metric}
\subsection{Analysis on Ranking Thickness and Existing Metric}
\label{sec:relation_thickness_other_metric}
% As shown in Table \ref{tab:comprehensive_result},
% R2ET and its variants are the winners on almost all the datasets except COMPAS and BP.
% Furthermore, on Adult and Bank,
% $\textnormal{R2ET}_{\backslash H}$ and R2ET-$\textnormal{mm}_{\backslash H}$
% that do not minimize Hessian norm
% outperform their counterparts (R2ET and R2ET-mm).
% We aim to identify the essential factor that helps a method outperform others.
As shown in Table \ref{tab:comprehensive_result},
R2ET and its variants are the winners on almost all the datasets except COMPAS and BP,
and
$\textnormal{R2ET}_{\backslash H}$ and R2ET-$\textnormal{mm}_{\backslash H}$
outperform their counterparts (R2ET and R2ET-mm) on Adult and Bank.
We aim to identify the essential factor that helps a method outperform others.
We will use the number of iterations to reach a successful attack where \textit{any} salient feature is swapped out of the top $k$ positions of the original explanation.
% and record the number of iterations to reach a successful attack.

\begin{figure}
    \centering
    \begin{minipage}{.24\textwidth}
    \includegraphics[width=\textwidth]{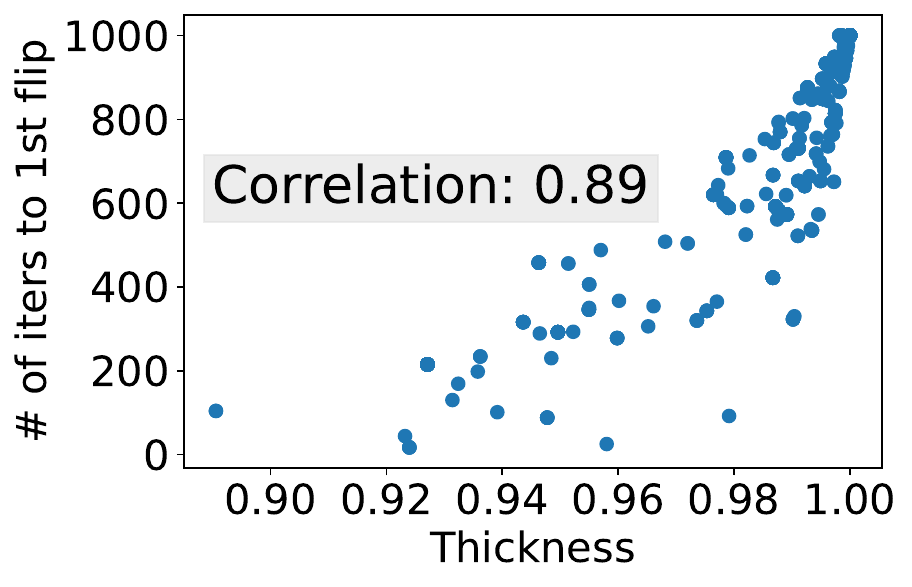}
    % \hspace{.25cm}%
    \end{minipage}%
    % \begin{minipage}{.2\textwidth}
    % \includegraphics[width=\textwidth]{figs/compas_$k$-IR and Manipulation epoch-no-y.pdf}
    % % \hspace{.25cm}%
    % \end{minipage}%
    % \begin{minipage}{.2\textwidth}
    % \includegraphics[width=\textwidth]{figs/compas_MSE and manipulation epoch-no-y.pdf}
    % % \hspace{.25cm}%
    % \end{minipage}%
    % \begin{minipage}{.2\textwidth}
    % \includegraphics[width=\textwidth]{figs/compas_PCC and manipulation epoch-no-y.pdf}
    % \end{minipage}%
    \begin{minipage}{.24\textwidth}
    \includegraphics[width=\textwidth]{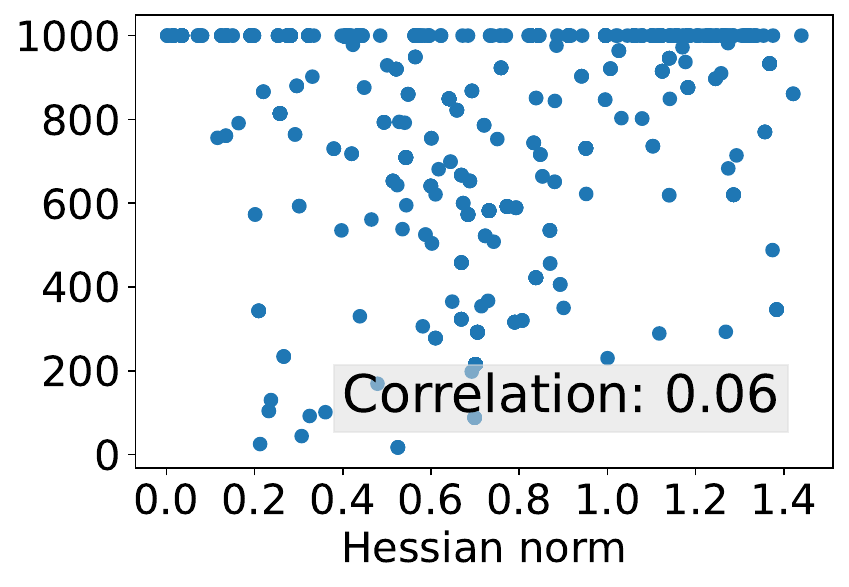}
    \end{minipage}%

    \caption{\small Correlation between the number of iterations to first flip, and ranking thickness (left)
    % P@$k$, MSE, PCC, 
    and Hessian norm (right) for R2ET model on COMPAS. 
    % More results are in Appendix \ref{sec:correlation_experiment_appendix}.
    % \textcolor{red}{Should be moved to the next page along with Sec. 5.3.}
    }
    \label{fig:correlation_manipulation_epoch}
    \vspace{-0.1in}
\end{figure}

% We run R2ET on COMPAS and BP datasets, record the Hessian norms at each input instance $\bx\in \mathcal{X}$,
% and then run ERAttack and 
% record the manipulation epoch,
% and $h(\bx, i, j)$ with $\bx^\prime$ set to the success attacks.
% and popular existing metrics,
% such as P@$k$, MSE, PCC, 
% and Hessian norm at the first successful attack.
To explore the correlations among metrics, 
for each sample $\bx$,
we collect the number of iterations to the first flip under ERAttack, Hessian norm, and thickness.
% estimated with $\bx^\prime$ being the adversarial samples.
Each dot in Fig. \ref{fig:correlation_manipulation_epoch} indicates one sample $\bx$,
% We plot these metrics
and the correlation coefficient between metrics is shown in each subplot.
% the manipulation epoch and thickness or Hessian norm
% is shown in the title of the subfigures.
% Apparently,
Compared with Hessian norm,
thickness is significantly more correlated with the number of iterations to the first flip.
% that is 0.89 v.s. 0.06 on COMPAS and 0.50 v.s. 0.13 on BP dataset.
It indicates that Hessian norm is \textit{not} a strong indicator of explanation ranking robustness.
% The thickness has the highest correlation (0.89) with manipulation epoch, and the best measurement to evaluate the model's explanation ranking robustness. It is worth noticing that none of P@$k$, MSE and PCC evaluate the robustness,
% but the similarity between two rankings.
% The Hessian norm is associated with the explanation robustness \cite{dombrowski2021towards,wang2020smoothed},
% but not the explanation \textit{ranking} robustness.
Recall the discussion in Sec \ref{sec:ranking_thickness_defense},
minimizing Hessian norm helps tighten the bounds. 
However, optimizing Hessian norm \textit{solely} 
% \cite{dombrowski2021towards} 
may only marginally contribute to the ranking robustness,
and it is consistent with the observations in Table \ref{tab:comprehensive_result},
where Est-H and Exact-H do not perform well most of the time.
As thickness is a precise measurement for the robustness of explanations, we further report the \textit{model-level} thickness, an average of thickness over all samples, in Table \ref{tab:thickess_patk}.
It could answer why R2ET cannot always outperform baselines:
On Adult and COMPAS, $\textnormal{R2ET}_{\backslash H}$ has a higher thickness (0.999 and 0.973, resp.) than R2ET (0.997 and 0.987, resp.).
We have similar observations for BP,
where Est-H has a higher thickness (0.93563) than R2ET-mm (0.93561).

% In sum, 
% models with higher thickness have better robustness performance, and R2ET cannot always reach the highest thickness and thus fail to outperform others.

\begin{table*}[h]
\caption{\small \textbf{P}@\boldsymbol{$k$} (shown in percentage) of different robust models (rows) under ERAttack and \textbf{model-level 
 thickness}.}
\vspace{-1pt}
\centering
\small
\begin{tabular}{ c || c c c | c | c c}
\toprule
\textbf{Method} & Adult & Bank
& COMPAS & MNIST & ADHD & BP \\
\midrule
Vanilla & 
87.6 / 0.9889 & 83.0 / 0.9692  & 84.2 / 0.9533 &
59.0 / 0.9725 
& 45.5 / 0.9261 
& 69.4 / 0.9282 
\\
WD &
91.7 / 0.9960  & 82.4 / 0.9568 & 87.7 / 0.9769 &
59.1 / 0.9732 
& 47.6 / 0.9343 
& 69.4 / 0.9298 
\\
SP & 
\underline{97.4} / \underline{0.9983} & 95.4 / \underline{0.9978} & \textbf{99.5} / \textbf{0.9999} &
62.9 / 0.9771 
& 42.5 / 0.9316
& 68.7 / 0.9300 
\\
Est-H & 
87.1 / 0.9875 & 78.4 / 0.9583 & 82.6 / 0.9557  &
85.2 / \underline{0.9948} 
& 58.2 / 0.9578
& \textbf{75.0} / \textbf{0.9356} 
\\
Exact-H & 
89.6 / 0.9932 & 81.9 / 0.9521 & 77.2 / 0.9382 &
- / - & - / - & - / -
\\
SSR & 91.2 / 0.9934 & 76.3 / 0.9370 & 82.1 / 0.9549 &
- / - & - / - & - / - 
\\
AT & 68.4 / 0.9372 & 80.0 / 0.9473 &  84.2 / 0.9168 & 56.0 / 0.9639
& 59.4 / 0.9597
& 72.0 / \underline{0.9342} 
\\
\midrule
$\textnormal{R2ET}_{\backslash H}$ & 
\textbf{97.5} / \textbf{0.9989}  & \textbf{100.0} / \textbf{1.0000} & 91.0 / 0.9727 &
82.8 / \textbf{0.9949} 
& 60.7 / 0.9588
& 70.9 / 0.9271 
\\
R2ET-$\textnormal{mm}_{\backslash H}$ &
93.5 / 0.9963 & \underline{95.8} / 0.9874 & \underline{95.3} / \underline{0.9906} &
81.6 / 0.9942 
& \underline{64.2} / \underline{0.9622}
& 72.4 / \underline{0.9342} 
\\
\midrule
R2ET & 
92.1 / 0.9970 & 80.4 / 0.9344 & 92.0 / 0.9865 &
\textbf{85.7} / \textbf{0.9949} 
& \textbf{71.6} / \textbf{0.9731}
 & 71.5 / 0.9296 
\\
R2ET-mm &
87.8 / 0.9943 & 75.1 / 0.9102 & 82.1 / 0.9544 &
\underline{85.3} / \underline{0.9948} 
& 58.8 / 0.9588
& \underline{73.8} / \textbf{0.9356} 
\\
\bottomrule
\end{tabular}
\label{tab:thickess_patk}
\end{table*}

\subsection{Case study: saliency maps visualization}
\label{sec:case_study}

\begin{figure*}[t]
     \centering
    %  \begin{minipage}{.8\textwidth}
    % %  \centering
    %  \includegraphics[width=\textwidth]{figs/case_study33_subset.pdf}
    %  % \hspace{.25cm}%
    %  \end{minipage}%
    %  \\
    %  \begin{minipage}{\textwidth}
    %  \centering
    % %  \includegraphics[width=\textwidth]{figs/case_study38_subset.pdf}
    %  \includegraphics[width=.9\textwidth]{figs/case-study_cifar10.pdf}
    %  % \hspace{.25cm}%
    %  \end{minipage}%
    \includegraphics[width=1\textwidth]{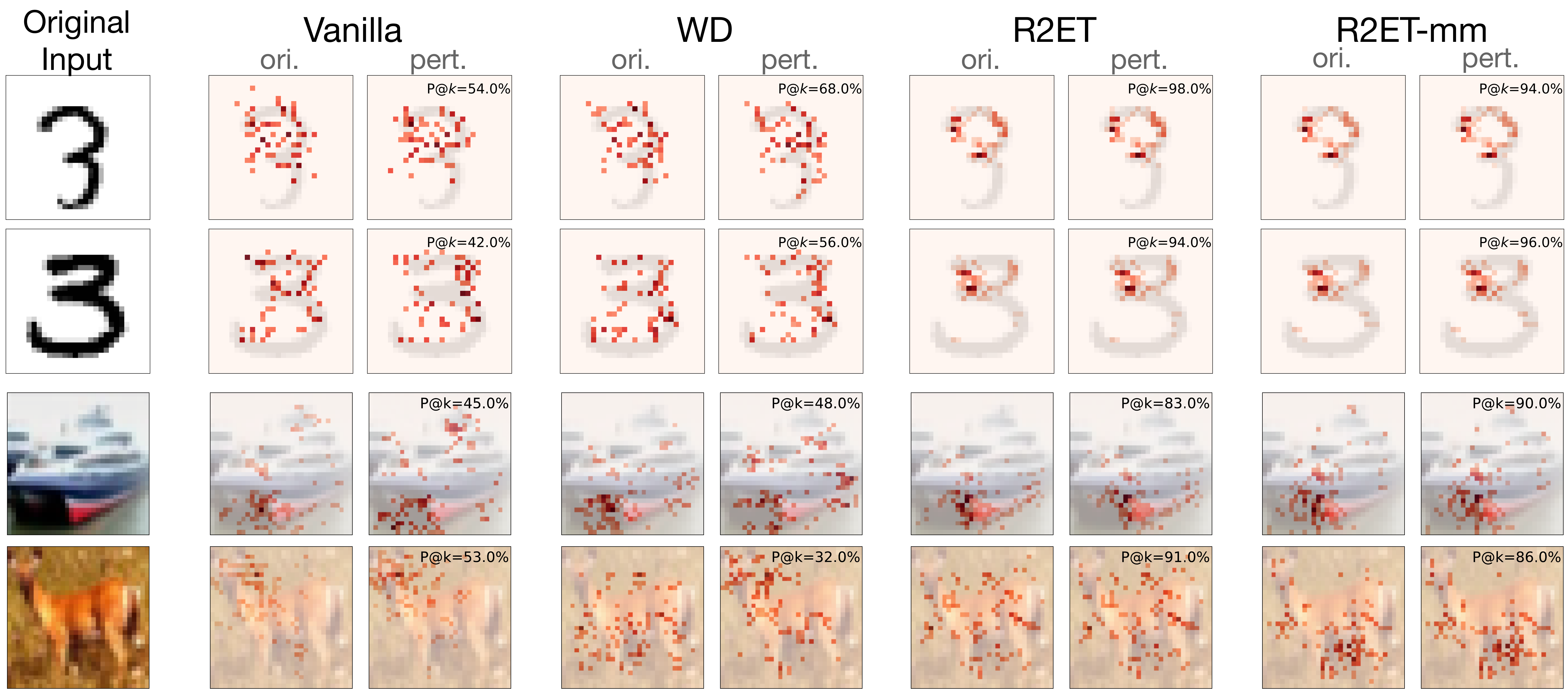}
     \caption{ \small
     \textbf{Top}: With the Siamese network that compares two images from MNIST,
     saliency maps for the original and perturbed pairs of input images under ERAttack against different robust methods. 
     The examples of a pair of images are from the same class (digit \textit{3}).
    %  The first two rows are examples of a pair of images of the same class (both are class 3), and the last two rows are images of different classes (classes 3 and 8). 
    \textbf{Bottom}: Two images from CIFAR-10, one from class \textit{ship} and another from class \textit{deer}.
    The top $k$ (50 for MNIST and 100 for CIFAR-10) salient pixels are highlighted, and darker colors indicate higher importance. 
    The robustness metric P@$k$ is printed within each subplot.
    % \textcolor{red}{The ordering of the CIFAR10 images is strange: the two input images are ordered vertically but the explanations of the two images are ordered horizontally.
    % }
    }
    \label{fig:case_study}
 \end{figure*}
% As shown in Table \ref{tab:comprehensive_result}, 
% R2ET and its variants are more effective against ERAttack and MSE attacks, 
% demonstrating that they can improve the ranking explanations robustness. 
% R2ET and its variants improve the ranking explanations robustness and are more effective against attacks.
Besides the robustness of ranking explanations, we hope optimizing thickness can lead to models that identify the ground-truth important features
(e.g., pixels covering the digits),
since a model that robustly uses irrelevant features is not useful in practice. 
% we explore how R2ET performs in terms of robustness and plausibility of explanations by visualizing samples on MNIST.
MNIST and CIFAR-10 provide suitable testbeds to visually evaluate whether R2ET forces the model to use irrelevant features.

In Fig.\ref{fig:case_study}, 
we conduct ERAttack against models trained with different strategies.
In MNIST, Vanilla performs poorly, with 
about 30$\%$ $\sim$ 50$\%$ of the top 50 important features fell out of top positions under ERAttack. 
% With weight decay (WD),
% % Using the model trained with weight decay, 
% P@$k$ is improved by about 15$\%$ compared to Vanilla in the example,
% but only 56$\%$ and 68$\%$.
WD improves P@$k$ 
% by about 15$\%$
to 56$\%$ and 68$\%$ compared with Vanilla in the example. 
For R2ET and R2ET-mm, 
it is difficult to find visible change in the top salient features before and after ERAttack, 
and P@$k$ of both methods are greater than 90$\%$. 
We can also see that 
the top 50 salient pixels used by Vanilla and WD do not highlight the spatial patterns of the digits.
However,
the explanations show that
the top 50 important features used by R2ET and R2ET-mm encode recognizable spatial patterns of the digits. 
We have similar conclusions in CIFAR-10.
Specifically,
take the explanations on the ship in Fig. \ref{fig:case_study} bottom as an example.
Vanilla and WD perform badly in terms of P@$k$, 
while R2ET and R2ET-mm achieve around as high as 90$\%$ P@$k$.
All four models make correct predictions due to a similar region
(front hull of the ship).
However, 
ERAttack manipulates the explanations of Vanilla and WD to include another region (wheelhouse of the ship)
while the key regions of the explanations of R2ET and R2ET-mm under attacks remain the same.
The wheelhouse may be one reason for classifying the image to ship, 
but the inconsistency of explanations due to imperceptible perturbations raises confusion and mistrust.
% See Fig. \ref{fig:case_study_all} for more results. 

% \subsection{Explanation Ranking Attacks and Constraints}
% \subsection{Sensitivity Analysis}

\subsection{Explanation Faithfulness}
\label{sec:experiment_faithfulness}
To ensure the \textit{faithfulness} of explanations for R2ET models, we evaluate their faithfulness using three widely accepted metrics. The results, presented in Table \ref{tab:faithfulness}, demonstrate that the explanations provided by R2ET models are on par with, if not superior to, other models in terms of faithfulness.

\begin{table*}[!htb]
\setlength{\tabcolsep}{4pt}
\caption{Faithfulness of explanations evaluated by DFFOT ($\downarrow$) / COMP ($\uparrow$) / SUFF ($\downarrow$).}
\centering
\small
\begin{tabular}{ c || c c c | c | c c}
\toprule
\textbf{Method} & Adult & Bank
& COMPAS & MNIST & ADHD & BP \\
\midrule
Vanilla 
& 0.24 / 0.43 / 0.18
& 0.23 / 0.14 / 0.04
& \textbf{0.17} / 0.37 / \underline{0.14} 
& 0.37 / 0.16 / 0.23
& 0.51 / 0.05 / 0.28
& 0.40 / \underline{0.06} / 0.29
\\
WD 
& 0.45 / 0.47 / 0.23
& 0.36 / 0.27 / 0.07 
& 0.29 / 0.41 / 0.18
& 0.37 / 0.16 / 0.22
& 0.49 / 0.06 / 0.27
& \textbf{0.35} / 0.05 / 0.33
\\
SP 
& 0.43 / 0.47 / 0.25
& 0.35 / 0.31 / 0.07 
& 0.29 / \textbf{0.45} / 0.18
& 0.38 / 0.15 / 0.22
& \textbf{0.30} / 0.10 / 0.34
& \underline{0.38} / \underline{0.06} / 0.30
\\
Est-H 
& 0.44 / 0.44 / 0.24
& 0.18 / 0.21 / 0.06
& 0.27 / 0.42 / 0.17
& 0.23 / \underline{0.24} / \textbf{0.18}
& 0.59 / 0.04 / \underline{0.26}
& 0.45 / 0.05 / \textbf{0.24}
\\
Exact-H 
& 0.43 / 0.46 / 0.23
& 0.19 / 0.14 / 0.04
& 0.30 / 0.40 / 0.18
& - / - / -
& - / - / -
& - / - / -
\\
SSR
& 0.54 / 0.39 / 0.21
& 0.46 / 0.04 / \textbf{0.01}
& 0.32 / 0.43 / 0.18
& - / - / -
& - / - / -
& - / - / -
\\
AT 
& \underline{0.16} / 0.14 / \textbf{0.08}
& 0.19 / 0.10 / \underline{0.03}
& \underline{0.24} / 0.10 / \textbf{0.07} 
& 0.40 / 0.12 / 0.28
& \underline{0.35} / 0.10 / \underline{0.26}
& 0.46 / \underline{0.06} / \underline{0.25}
\\
\midrule
$\textnormal{R2ET}_{\backslash H}$ 
& \textbf{0.13} / \textbf{0.50} / \underline{0.14}
& 0.34 / \underline{0.32} / 0.10
& \textbf{0.17} / 0.40 / 0.17
& 0.23 / 0.22 / \underline{0.19}
& 0.38 / \underline{0.13} / 0.37
& 0.43 / \textbf{0.07} / 0.29
\\
R2ET-$\textnormal{mm}_{\backslash H}$ 
& 0.42 / 0.47 / 0.22
& 0.34 / \textbf{0.41} / 0.14
& 0.25 / 0.42 / 0.17
& 0.25 / 0.22 / 0.21
& 0.37 / \textbf{0.17} / 0.37
& 0.42 / \textbf{0.07} / 0.29
\\
\midrule
R2ET 
& 0.32 / 0.46 / 0.19
& \textbf{0.11} / 0.24 / 0.07
& 0.27 / 0.39 / 0.17
& \textbf{0.18} / \textbf{0.26} / 0.23
& 0.48 / 0.12 / \underline{0.26}
& 0.42 / \textbf{0.07} / 0.29
\\
R2ET-mm 
& 0.38 / \underline{0.48} / 0.20
& \underline{0.12} / 0.21 / 0.08
& 0.28 / \underline{0.44} / 0.15
& \underline{0.19} / \textbf{0.26} / 0.22
& 0.50 / 0.04 / \textbf{0.25}
& 0.45 / 0.05 / 0.29

\\
\bottomrule
\end{tabular}
\label{tab:faithfulness}
\end{table*}

\subsection{Sensitivity Analysis}

\begin{table}[!htb]
\caption{P@$k$ (shown in percentage) of models under ERAttack when the models are trained from a \textbf{random state} or \textbf{retrained} from the Vanilla models.}
\small
\centering
\begin{tabular}{ c || c c c}
\toprule
\textbf{Method} & Adult & Bank
& COMPAS \\
\midrule
Vanilla & 
87.6 / 87.6 & 
83.0 / 83.0 & 
84.2 / 84.2
\\
WD &
91.7 / 88.3 & 
82.4 / 82.1 & 
87.7 / 82.7
\\
% CL \cite{hein2017formal} & 
CL &
- / 93.1 & 
- / 100.0 & 
- / 87.1
\\
Est-H & 
87.1 / 92.1 & 
78.4 / 85.2 & 
82.6 / 85.1
\\
Exact-H & 
89.6 / 88.7 & 
81.9 / - & 
77.2 / 87.0
\\
SSR & 
91.2 / 88.7 & 
76.3 / - & 
82.1 / 86.1
\\
\midrule
$\textnormal{R2ET}_{\backslash H}$ & 
97.5 / 100.0 & 
100.0 / 100.0 & 
91.9 / 97.8
\\
R2ET-$\textnormal{mm}_{\backslash H}$ &
93.5 / 100.0 & 
95.8 / 98.3 &
95.3 / 95.6
\\
\midrule
R2ET & 
92.1 / 92.6 & 
80.4 / 86.2 & 
92.0 / 85.1
\\
R2ET-mm &
87.8 / 91.6 &
75.1 / 86.2 & 
82.1 / 87.4
\\
\bottomrule
\end{tabular}
\label{tab:sensitivity_analysis_pretrain}
\end{table}

\noindent \textbf{Impacts of pretrain / retrain.}
% Lastly,
We explore how good are these methods when applying them in the \textbf{retrain} schema.
In previous experiments, 
all models are trained from random states.
We now retrain the Vanilla models with these methods for 10 epochs at most.
Since the Vanilla model has already converged and reached a good cAUC,
we assume that the Vanilla model's explanation ranking is an excellent reference,
and thus these robust methods try to maintain the Vanilla model's rankings.
Thus, we will terminate the retraining phase if
P@$k$ between Vanilla model's explanation ranking and the retrain model's ranking significantly drops,
or the retrain model's cAUC drops a lot.

Table \ref{tab:sensitivity_analysis_pretrain} presents the results for comparing two training schemas.
Since the baseline SP changes the models' structure (activation function),
we do not consider it here.
Instead, 
we consider another baseline, 
CL \cite{hein2017formal},
and it is adopted here due to much fewer re-training epochs.
More details for CL can be found in Sec. \ref{sec:ranking_thickness_defense}.
Besides that,
\textit{none} of retrain models by Exact-H and SSR can maintain Vanilla model's explanation rankings and cAUC at the same time on Bank,
and thus both are not applicable.

\section{Related Work}
\label{sec:related_work}

\noindent \textbf{Explainable machine learning and explanation robustness.}
Recent post-hoc explanation methods for deep networks can be categorized into gradient-based \cite{zhou2016learning,selvaraju2017grad,baldassarre2019explainability,smilkov2017smoothgrad,sundararajan2017axiomatic,shrikumar2017learning}, 
surrogate model based \cite{ribeiro2016should,huang2022graphlime}, 
Shapley values \cite{lundberg2017unified,liu2020shapley,ancona2019explaining}, 
and causality \cite{pearl2018theoretical,chattopadhyay2019neural}.
% for tabular data \cite{rudin2023globally}, images \cite{selvaraju2017grad,ribeiro2016should} and graphs \cite{ying2019gnnexplainer,baldassarre2019explainability}.
% 
% \cite{chen2021self,dhurandhar2018explanations,selvaraju2017grad,smilkov2017smoothgrad}.
% There are sparse work for the explanations of SN \cite{chen2021self,faber2020contrastive,lanciano2020explainable,plummer2020these}.
% \noindent \textbf{Robust Explanations.}
Although gradient-based methods are widely used \cite{nielsen2022robust}, 
they are found to lack robustness against small perturbations \cite{ghorbani2019interpretation,heo2019fooling}.
% The gradient-based methods are widely used in practice due to their simplicity and efficiency \cite{nielsen2022robust}.
% However,
% existing works found that gradient-based explanations are not robust to small perturbations \cite{ghorbani2019interpretation,heo2019fooling}.
% Authors in \cite{ghorbani2019interpretation} conduct adversarial attacks on explanations to explore the sensitivity of explanations.
Some works \cite{chen2019robust,singh2020attributional,ivankay2020far,wang2022exploiting,sarkar2021enhanced,upadhyay2021towards} propose to improve the explanation robustness by adversarial training (AT).
% To improve the explanation robustness,
To bypass the high time complexity of AT,
some works propose 
% other training strategies, 
% such as
replacing ReLU function with softplus \cite{dombrowski2021towards},
training with weight decay 
% to improve the robustness of explanations
\cite{dombrowski2019explanations},
% Authors in \cite{dombrowski2021towards,wang2020smoothed} propose to
and
incorporating gradient- and Hessian-related terms as regularizers \cite{wang2020smoothed,wicker2022robust}.
% In \cite{alvarez2018robustness,alvarez2018towards,wang2020smoothed},
% Authors in \cite{alvarez2018towards,wang2020smoothed}
% derive the definition of explanation robustness based on Lipschitz. 
% There is few existing work on explanation ranking robustness.
% Authors in \cite{wang2022exploiting} connect Kendall's Rank Correlation and Cosine similarity for explanations.
% Instead of proposing \textit{training} methods, 
Some works propose \textit{explanation} methods, 
rather than \textit{training} methods,
to enhance explanation robustness \cite{smilkov2017smoothgrad,lu2021dance,liu2022certifiably,chen2021self,manupriya2022improving,rieger2020simple}.
Besides, many works \cite{madry2017towards,tu2019theoretical,roth2020adversarial,yang2020boundary,yang2020closer,deng2021improving,tsipras2018robustness,wen2020towards,zhang2019theoretically} for adversarial robustness focus on \textit{prediction} robustness, instead of \textit{explanation} robustness.

\noindent \textbf{Ranking robustness and manipulations.}
The ranking robustness is well-studied in information retrieval (IR),
in terms of ``noise'' \cite{zhou2006ranking} and adversarial attacks \cite{goren2018ranking}.
% Authors in \cite{zhou2006ranking} 
% work on the ranking robustness in terms of ``noise''.
% formulate ranking robustness as the expectation similarity between rankings with and without noise.
% In \cite{goren2018ranking},
% authors explore the robustness under adversarial attacks.
% and define Lipschitz-related robustness.
% Authors in \cite{wang2012robust} propose to train a robust ranking model by risk-sensitive optimization. 
% In \cite{zhou2020adversarial,zhou2021practical},
% authors consider attacks on either query or the candidate set to manipulate the ranking within one or more pairs of images, respectively.
% Authors in \cite{wang2020transferable} maximize the similarity of mismatched image pairs and minimize the similarity of matched image pairs simultaneously to manipulate the ranking.
% 
% 
% 
% \noindent \textbf{Attacking ranking algorithms.}
% %
% Authors in \cite{ghorbani2019interpretation} propose a ``top-$k$'' attack over explanations,
% % where $h(\mathbf{x},i,j)=\mathcal{I}(\mathbf{x})_i$.
% % Unfortunately,
% % it may result in an unexpected solution where all, 
% % not only the top-$k$ ones,
% % of the gradient magnitudes diminish while the original explanation ranking is reserved.
% which minimizes the magnitude of the top-$k$ features and ignores the rest ones.
% %
% For example,
% $\mathcal{I}(\mathbf{x})_i$ reduces to $0.01*\mathcal{I}(\mathbf{x})_i$, 
% $\forall i$.
% On the contrary, 
% conducting ERAttack based on Eq. (\ref{eq:attack_thick}) inhibits the top-$k$ and promotes the rest,
% thus avoiding the trivial solution that minimizes all gradient magnitudes.
In \cite{wang2020transferable},
% from image retrieval area,
authors attacked image ranking by maximizing (minimizing, resp.) the similarity of mismatched (matched resp.) image pairs.
% The attackers either focus on only one pair, 
% % \cite{zhou2020adversarial},
% or distort multiple rankings by manipulating the ``query''
% % \cite{li2021qair,zhou2020adversarial,zhou2021practical}.
% \cite{zhou2020adversarial,zhou2021practical}.
% Besides that, 
% these works \cite{wang2020transferable,zhou2020adversarial,zhou2021practical} 
% apply a hinge loss over the difference between two images, 
% following the triplet loss widely used in Contrastive Learning \cite{ma2019deep}.
% However, 
% we do not use the hinge loss in our task due to the high correlation among elements in the ranking,
% where elements are different entries of the gradient of the \textit{same} model with respect to the \textit{same} input,
% and every entry can be dependent on each other.
Black-box attacks \cite{li2021qair} and targeted manipulations \cite{tolias2019targeted} on rankings are also studied.
% In image retrieval, 
% elements in the ranking are independent individual images to be selected.
% However,
% elements in our ranking are different entries of the gradient of the \textit{same} model with respect to the \textit{same} input,
% and every entry can be dependent on each other.
% Thus,
% discarding the loss from attack-favorable cases, 
% e.g., $h(\mathbf{x},i,j)<0$,
% is more likely to result in the fluctuation of $\delta$ update.
% Take Fig. \ref{fig:example_attack_direction} as an example,
% $\delta$ moves in the optimal direction when considering all the cases during attacks, 
% shown in the left two sub-figures.
% However,
% $\delta$ will be trapped and confused about the true update direction if it considers the $(i,j)$ pair with $h(\mathbf{x},i,j)>0$ cases only.
% % [For example, $\mathcal{I}=[2x_1-x_2,x_2-2x_3,2x_3-x_1]^\top$ where $x_i$ is the $i$-th entry of input.
% % In the case of $h(\mathbf{x}^\prime,1,2)=2x_1-2x_3>0$ and $h(\mathbf{x}^\prime,2,3)=x_2-x_1<0$,
% % $\delta$ will be updated to decrease $x_1$ and increase $x_3$ to minimize the Eq. (\ref{eq:attack_thick}).
% % However, the decreasing of $x_1$ may make $(\mathbf{x}^\prime,1,2)<0$ and $h(\mathbf{x}^\prime,2,3)>0$,
% % and the direction to update $\delta$ converts to increase $x_1$.
% % Therefore, 
% % $\delta$ will be trapped and confused about the true update direction with respect to $x_1$.
% % ]
Attacks on IR and explanations are different in three aspects.
1) Vulnerabilities: IR has queries and candidates that can be attacked \cite{zhou2020adversarial,zhou2021practical}, while we focus on attacking the gradient of classifiers via input manipulation.
2) Attacking objectives: On IR, authors either manipulate the ranking of \textit{one} single candidate, or manipulate query to distort the ranking of candidates. 
We aim to swap \textit{any} pairs of salient and non-salient features.
% and we do not find a similar goal in IR papers.
3) Computations: 
explanations are defined by gradient or its variants and studying their robustness requires second or higher-order derivatives. 
The cost of higher-order derivatives motivates us to design a surrogate regularizer to bypass the costly computations.

% \noindent \textbf{Hessian Approximations.}
% Multiple works study the approximation of Hessian matrix, 
% which can be roughly divided into the following three groups.
% First, 
% authors in \cite{pearlmutter1994fast,moosavi2019robustness} use a finite differentiation form to approximate the Hessian matrix,
% which also known as ``Pearlmutter's trick''.
% Second,
% authors in \cite{dombrowski2021towards} estimate the expectation of the product between Hessian matrix and a zero-mean Gaussian vector by Monte Carlo.
% Third,
% in \cite{dombrowski2021towards},
% the Hessian matrix is approximated by computing the inner product of a random vector with the first-order gradient.
% In \cite{wang2020smoothed},
% they approximate the Hessian matrix by a closed-form formula given a specific group of models.
% Instead of focusing on Hessian matrix,
% authors in \cite{ghorbani2019investigation} investigate the impacts of the eigenvalues of Hessian matrix in neural networks, 
% and they use a stochastic Lanczos quadrature algorithm to estimate the Hessian density.

\section{Conclusion}
\label{sec:conclusion}
We proposed ``\textit{explanation ranking thickness}'' to measure the robustness of the top-ranked salient explaining features to align with human cognitive capability when interpreting a classifier's predictions.
We provided theoretical insights, including surrogate bounds of the thickness, the connection between thickness and a min-max optimization problem, and a global convergence rate of a constrained multi-objective attacking algorithm against the thickness.
The theory leads to a well-justified optimization problem and an efficient training algorithm \textit{R2ET}.
On 7 datasets (vectors, images, and graphs) and with 2 neural network architectures, we compared 7 state-of-the-art baselines and 3 variants of R2ET, and consistently confirmed that explanation ranking thickness is indeed a strong indicator of the stability of top salient features.
In the future, we plan to consider it in the natural language processing problems and to explore R2ET with larger language models.
% We define the explanation ranking thickness for a pair of and top-$k$ features, respectively.
% Then we propose a ranking-based explanation attack, 
% ERAttack,
% to effectively manipulate the explanation rankings compared with existing MSE attack. 
% To avoid the expensive neighbor sampling and adversarial training,
% we propose R2ET,
% a ranking-specific defense method,
% which adopts the surrogate bounds of the thickness as regularizer,
% to efficiently train ranking robust models.
% We further confirm that the thickness can be measured by the proposed thickness metric by analyzing the convergence of the gradient-based attacks and experimental results.
% to alleviate the attacks by improving the lower bounds.
% Even though we show the close correlation between thickness and explanation robustness,
% we cannot directly optimize the thickness as part of objective functions for robust training.
% % Instead,
% % we have to relax the non-differentiable operations,
% % such as indicator functions,
% % and use proxy loss functions.
% Consequently,
% R2ET does not necessarily have the highest thickness,
% thus cannot consistently outperforms other baselines.

\section*{Acknowledgements}
Chao Chen and Sihong Xie were supported in part by the National Science Foundation under NSF Grants IIS-1909879, CNS-1931042, IIS-2008155, and IIS-2145922.
Chenghua Guo and Xi Zhang were supported by the Natural Science Foundation of China (No. 61976026) and the CAAI-Huawei MindSpore Open Fund.

\clearpage
\newpage
\bibliography{icml2022.bib}

\begin{thebibliography}{69}
\providecommand{\natexlab}[1]{#1}
\providecommand{\url}[1]{\texttt{#1}}
\expandafter\ifx\csname urlstyle\endcsname\relax
  \providecommand{\doi}[1]{doi: #1}\else
  \providecommand{\doi}{doi: \begingroup \urlstyle{rm}\Url}\fi

\bibitem[Adebayo et~al.(2018)Adebayo, Gilmer, Muelly, Goodfellow, Hardt, and
  Kim]{adebayo2018sanity}
Adebayo, J., Gilmer, J., Muelly, M., Goodfellow, I., Hardt, M., and Kim, B.
\newblock Sanity checks for saliency maps.
\newblock \emph{NeurIPS}, 31, 2018.

\bibitem[Ancona et~al.(2019)Ancona, Oztireli, and Gross]{ancona2019explaining}
Ancona, M., Oztireli, C., and Gross, M.
\newblock Explaining deep neural networks with a polynomial time algorithm for
  shapley value approximation.
\newblock In \emph{ICML}, 2019.

\bibitem[Baldassarre \& Azizpour(2019)Baldassarre and
  Azizpour]{baldassarre2019explainability}
Baldassarre, F. and Azizpour, H.
\newblock Explainability techniques for graph convolutional networks.
\newblock \emph{ICML}, 2019.

\bibitem[Chattopadhyay et~al.(2019)Chattopadhyay, Manupriya, Sarkar, and
  Balasubramanian]{chattopadhyay2019neural}
Chattopadhyay, A., Manupriya, P., Sarkar, A., and Balasubramanian, V.~N.
\newblock Neural network attributions: A causal perspective.
\newblock In \emph{ICML}, 2019.

\bibitem[Chen et~al.(2021)Chen, Shen, Ma, Kong, Rangarajan, Zhang, and
  Xie]{chen2021self}
Chen, C., Shen, Y., Ma, G., Kong, X., Rangarajan, S., Zhang, X., and Xie, S.
\newblock Self-learn to explain siamese networks robustly.
\newblock \emph{ICDM}, 2021.

\bibitem[Chen et~al.(2019)Chen, Wu, Rastogi, Liang, and Jha]{chen2019robust}
Chen, J., Wu, X., Rastogi, V., Liang, Y., and Jha, S.
\newblock Robust attribution regularization.
\newblock \emph{NeurIPS}, 2019.

\bibitem[Deng et~al.(2021)Deng, Zhang, Ghorbani, and Zou]{deng2021improving}
Deng, Z., Zhang, L., Ghorbani, A., and Zou, J.
\newblock Improving adversarial robustness via unlabeled out-of-domain data.
\newblock In \emph{AISTATS}, 2021.

\bibitem[DeYoung et~al.(2019)DeYoung, Jain, Rajani, Lehman, Xiong, Socher, and
  Wallace]{deyoung2019eraser}
DeYoung, J., Jain, S., Rajani, N.~F., Lehman, E., Xiong, C., Socher, R., and
  Wallace, B.~C.
\newblock Eraser: A benchmark to evaluate rationalized nlp models.
\newblock \emph{ACL}, 2019.

\bibitem[Dombrowski et~al.(2019)Dombrowski, Alber, Anders, Ackermann,
  M{\"u}ller, and Kessel]{dombrowski2019explanations}
Dombrowski, A.-K., Alber, M., Anders, C.~J., Ackermann, M., M{\"u}ller, K.-R.,
  and Kessel, P.
\newblock Explanations can be manipulated and geometry is to blame.
\newblock \emph{NeurIPS}, 2019.

\bibitem[Dombrowski et~al.(2021)Dombrowski, Anders, M{\"u}ller, and
  Kessel]{dombrowski2021towards}
Dombrowski, A.-K., Anders, C.~J., M{\"u}ller, K.-R., and Kessel, P.
\newblock Towards robust explanations for deep neural networks.
\newblock \emph{Pattern Recognition}, 2021.

\bibitem[Ghorbani et~al.(2019)Ghorbani, Abid, and
  Zou]{ghorbani2019interpretation}
Ghorbani, A., Abid, A., and Zou, J.
\newblock Interpretation of neural networks is fragile.
\newblock In \emph{AAAI}, 2019.

\bibitem[Goodman \& Flaxman(2017)Goodman and Flaxman]{goodman2017european}
Goodman, B. and Flaxman, S.
\newblock European union regulations on algorithmic decision-making and a
  “right to explanation”.
\newblock \emph{AI magazine}, 2017.

\bibitem[Goren et~al.(2018)Goren, Kurland, Tennenholtz, and
  Raiber]{goren2018ranking}
Goren, G., Kurland, O., Tennenholtz, M., and Raiber, F.
\newblock Ranking robustness under adversarial document manipulations.
\newblock In \emph{SIGIR}, 2018.

\bibitem[Hein \& Andriushchenko(2017)Hein and Andriushchenko]{hein2017formal}
Hein, M. and Andriushchenko, M.
\newblock Formal guarantees on the robustness of a classifier against
  adversarial manipulation.
\newblock \emph{NeurIPS}, 30, 2017.

\bibitem[Heo et~al.(2019)Heo, Joo, and Moon]{heo2019fooling}
Heo, J., Joo, S., and Moon, T.
\newblock Fooling neural network interpretations via adversarial model
  manipulation.
\newblock \emph{NeurIPS}, 2019.

\bibitem[Huang et~al.(2022)Huang, Yamada, Tian, Singh, and
  Chang]{huang2022graphlime}
Huang, Q., Yamada, M., Tian, Y., Singh, D., and Chang, Y.
\newblock Graphlime: Local interpretable model explanations for graph neural
  networks.
\newblock \emph{TKDE}, 2022.

\bibitem[Huang et~al.(2016)Huang, Xu, Schuurmans, and
  Szepesv{\'a}ri]{huang2015learning}
Huang, R., Xu, B., Schuurmans, D., and Szepesv{\'a}ri, C.
\newblock Learning with a strong adversary.
\newblock \emph{ICLR}, 2016.

\bibitem[Hudson \& Cohen(2000)Hudson and Cohen]{hudson2000neural}
Hudson, D.~L. and Cohen, M.~E.
\newblock \emph{Neural networks and artificial intelligence for biomedical
  engineering}.
\newblock Wiley Online Library, 2000.

\bibitem[Ivankay et~al.(2020)Ivankay, Girardi, Marchiori, and
  Frossard]{ivankay2020far}
Ivankay, A., Girardi, I., Marchiori, C., and Frossard, P.
\newblock Far: A general framework for attributional robustness.
\newblock \emph{BMVC}, 2020.

\bibitem[Krizhevsky et~al.(2009)Krizhevsky, Hinton,
  et~al.]{krizhevsky2009learning}
Krizhevsky, A., Hinton, G., et~al.
\newblock Learning multiple layers of features from tiny images.
\newblock 2009.

\bibitem[LeCun et~al.(1998)LeCun, Bottou, Bengio, and
  Haffner]{lecun1998gradient}
LeCun, Y., Bottou, L., Bengio, Y., and Haffner, P.
\newblock Gradient-based learning applied to document recognition.
\newblock \emph{Proceedings of the IEEE}, 1998.

\bibitem[Li et~al.(2020)Li, Wang, Jana, and Carin]{li2020towards}
Li, B., Wang, S., Jana, S., and Carin, L.
\newblock Towards understanding fast adversarial training.
\newblock \emph{arXiv preprint arXiv:2006.03089}, 2020.

\bibitem[Li et~al.(2021)Li, Li, Chen, Ye, He, Wang, Su, and Xue]{li2021qair}
Li, X., Li, J., Chen, Y., Ye, S., He, Y., Wang, S., Su, H., and Xue, H.
\newblock Qair: Practical query-efficient black-box attacks for image
  retrieval.
\newblock In \emph{CVPR}, 2021.

\bibitem[Liu et~al.(2022)Liu, Chen, Liu, Xia, and Gan]{liu2022certifiably}
Liu, A., Chen, X., Liu, S., Xia, L., and Gan, C.
\newblock Certifiably robust interpretation via r{\'e}nyi differential privacy.
\newblock \emph{Artificial Intelligence}, 2022.

\bibitem[Liu et~al.(2020)Liu, Chen, Liu, Zhang, and Xie]{liu2020shapley}
Liu, Y., Chen, C., Liu, Y., Zhang, X., and Xie, S.
\newblock Shapley values and meta-explanations for probabilistic graphical
  model inference.
\newblock In \emph{CIKM}, 2020.

\bibitem[Lu et~al.(2021)Lu, Guo, Xing, and Noble]{lu2021dance}
Lu, Y.~Y., Guo, W., Xing, X., and Noble, W.~S.
\newblock Dance: Enhancing saliency maps using decoys.
\newblock In \emph{ICML}, 2021.

\bibitem[Lundberg \& Lee(2017)Lundberg and Lee]{lundberg2017unified}
Lundberg, S.~M. and Lee, S.-I.
\newblock A unified approach to interpreting model predictions.
\newblock \emph{NeurIPS}, 2017.

\bibitem[Ma et~al.(2016)Ma, He, Cao, Zhang, Yu, and Ragin]{ma2016multi}
Ma, G., He, L., Cao, B., Zhang, J., Yu, P.~S., and Ragin, A.~B.
\newblock Multi-graph clustering based on interior-node topology with
  applications to brain networks.
\newblock In \emph{ECML PKDD}, 2016.

\bibitem[Ma et~al.(2019)Ma, Ahmed, Willke, Sengupta, Cole, Turk-Browne, and
  Yu]{ma2019deep}
Ma, G., Ahmed, N.~K., Willke, T.~L., Sengupta, D., Cole, M.~W., Turk-Browne,
  N.~B., and Yu, P.~S.
\newblock Deep graph similarity learning for brain data analysis.
\newblock In \emph{CIKM}, 2019.

\bibitem[Madry et~al.(2017)Madry, Makelov, Schmidt, Tsipras, and
  Vladu]{madry2017towards}
Madry, A., Makelov, A., Schmidt, L., Tsipras, D., and Vladu, A.
\newblock Towards deep learning models resistant to adversarial attacks.
\newblock \emph{ICLR}, 2017.

\bibitem[Manupriya et~al.(2022)Manupriya, Menta, Jagarlapudi, and
  Balasubramanian]{manupriya2022improving}
Manupriya, P., Menta, T.~R., Jagarlapudi, S.~N., and Balasubramanian, V.~N.
\newblock Improving attribution methods by learning submodular functions.
\newblock In \emph{AISTATS}, 2022.

\bibitem[Moosavi-Dezfooli et~al.(2019)Moosavi-Dezfooli, Fawzi, Uesato, and
  Frossard]{moosavi2019robustness}
Moosavi-Dezfooli, S.-M., Fawzi, A., Uesato, J., and Frossard, P.
\newblock Robustness via curvature regularization, and vice versa.
\newblock In \emph{CVPR}, 2019.

\bibitem[Moro et~al.(2014)Moro, Cortez, and Rita]{moro2014data}
Moro, S., Cortez, P., and Rita, P.
\newblock A data-driven approach to predict the success of bank telemarketing.
\newblock \emph{Decision Support Systems}, 2014.

\bibitem[Mothilal et~al.(2020)Mothilal, Sharma, and
  Tan]{mothilal2020explaining}
Mothilal, R.~K., Sharma, A., and Tan, C.
\newblock Explaining machine learning classifiers through diverse
  counterfactual explanations.
\newblock In \emph{FAccT}, 2020.

\bibitem[Nielsen et~al.(2022)Nielsen, Dera, Rasool, Ramachandran, and
  Bouaynaya]{nielsen2022robust}
Nielsen, I.~E., Dera, D., Rasool, G., Ramachandran, R.~P., and Bouaynaya, N.~C.
\newblock Robust explainability: A tutorial on gradient-based attribution
  methods for deep neural networks.
\newblock \emph{IEEE Signal Processing Magazine}, 2022.

\bibitem[Pearl(2018)]{pearl2018theoretical}
Pearl, J.
\newblock Theoretical impediments to machine learning with seven sparks from
  the causal revolution.
\newblock \emph{WSDM}, 2018.

\bibitem[Pearlmutter(1994)]{pearlmutter1994fast}
Pearlmutter, B.~A.
\newblock Fast exact multiplication by the hessian.
\newblock \emph{Neural computation}, 1994.

\bibitem[Pu \& Chen(2006)Pu and Chen]{pu2006trust}
Pu, P. and Chen, L.
\newblock Trust building with explanation interfaces.
\newblock In \emph{IUI}, 2006.

\bibitem[Ribeiro et~al.(2016)Ribeiro, Singh, and Guestrin]{ribeiro2016should}
Ribeiro, M.~T., Singh, S., and Guestrin, C.
\newblock " why should i trust you?" explaining the predictions of any
  classifier.
\newblock In \emph{SIGKDD}, 2016.

\bibitem[Rieger \& Hansen(2020)Rieger and Hansen]{rieger2020simple}
Rieger, L. and Hansen, L.~K.
\newblock A simple defense against adversarial attacks on heatmap explanations.
\newblock \emph{WHI}, 2020.

\bibitem[Roth et~al.(2020)Roth, Kilcher, and Hofmann]{roth2020adversarial}
Roth, K., Kilcher, Y., and Hofmann, T.
\newblock Adversarial training is a form of data-dependent operator norm
  regularization.
\newblock \emph{NeurIPS}, 2020.

\bibitem[Saaty \& Ozdemir(2003)Saaty and Ozdemir]{saaty2003magic}
Saaty, T.~L. and Ozdemir, M.~S.
\newblock Why the magic number seven plus or minus two.
\newblock \emph{Mathematical and computer modelling}, 2003.

\bibitem[Sarkar et~al.(2021)Sarkar, Sarkar, and
  Balasubramanian]{sarkar2021enhanced}
Sarkar, A., Sarkar, A., and Balasubramanian, V.~N.
\newblock Enhanced regularizers for attributional robustness.
\newblock In \emph{AAAI}, 2021.

\bibitem[Selvaraju et~al.(2017)Selvaraju, Cogswell, Das, Vedantam, Parikh, and
  Batra]{selvaraju2017grad}
Selvaraju, R.~R., Cogswell, M., Das, A., Vedantam, R., Parikh, D., and Batra,
  D.
\newblock Grad-cam: Visual explanations from deep networks via gradient-based
  localization.
\newblock In \emph{ICCV}, 2017.

\bibitem[Serrano \& Smith(2019)Serrano and Smith]{serrano2019attention}
Serrano, S. and Smith, N.~A.
\newblock Is attention interpretable?
\newblock \emph{ACL}, 2019.

\bibitem[Shrikumar et~al.(2017)Shrikumar, Greenside, and
  Kundaje]{shrikumar2017learning}
Shrikumar, A., Greenside, P., and Kundaje, A.
\newblock Learning important features through propagating activation
  differences.
\newblock In \emph{ICML}, 2017.

\bibitem[Singh et~al.(2020)Singh, Kumari, Mangla, Sinha, Balasubramanian, and
  Krishnamurthy]{singh2020attributional}
Singh, M., Kumari, N., Mangla, P., Sinha, A., Balasubramanian, V.~N., and
  Krishnamurthy, B.
\newblock Attributional robustness training using input-gradient spatial
  alignment.
\newblock In \emph{ECCV}, 2020.

\bibitem[Smilkov et~al.(2017)Smilkov, Thorat, Kim, Vi{\'e}gas, and
  Wattenberg]{smilkov2017smoothgrad}
Smilkov, D., Thorat, N., Kim, B., Vi{\'e}gas, F., and Wattenberg, M.
\newblock Smoothgrad: removing noise by adding noise.
\newblock \emph{arXiv preprint arXiv:1706.03825}, 2017.

\bibitem[Sundararajan et~al.(2017)Sundararajan, Taly, and
  Yan]{sundararajan2017axiomatic}
Sundararajan, M., Taly, A., and Yan, Q.
\newblock Axiomatic attribution for deep networks.
\newblock In \emph{ICML}, 2017.

\bibitem[Tolias et~al.(2019)Tolias, Radenovic, and Chum]{tolias2019targeted}
Tolias, G., Radenovic, F., and Chum, O.
\newblock Targeted mismatch adversarial attack: Query with a flower to retrieve
  the tower.
\newblock In \emph{ICCV}, 2019.

\bibitem[Tsipras et~al.(2018)Tsipras, Santurkar, Engstrom, Turner, and
  Madry]{tsipras2018robustness}
Tsipras, D., Santurkar, S., Engstrom, L., Turner, A., and Madry, A.
\newblock Robustness may be at odds with accuracy.
\newblock \emph{ICLR}, 2018.

\bibitem[Tu et~al.(2019)Tu, Zhang, and Tao]{tu2019theoretical}
Tu, Z., Zhang, J., and Tao, D.
\newblock Theoretical analysis of adversarial learning: A minimax approach.
\newblock \emph{NeurIPS}, 2019.

\bibitem[Upadhyay et~al.(2021)Upadhyay, Joshi, and
  Lakkaraju]{upadhyay2021towards}
Upadhyay, S., Joshi, S., and Lakkaraju, H.
\newblock Towards robust and reliable algorithmic recourse.
\newblock \emph{NeurIPS}, 2021.

\bibitem[Wang \& Kong(2022)Wang and Kong]{wang2022exploiting}
Wang, F. and Kong, A. W.-K.
\newblock Exploiting the relationship between kendall's rank correlation and
  cosine similarity for attribution protection.
\newblock \emph{NeurIPS}, 2022.

\bibitem[Wang et~al.(2020{\natexlab{a}})Wang, Wang, Li, Zhang, and
  Lin]{wang2020transferable}
Wang, H., Wang, G., Li, Y., Zhang, D., and Lin, L.
\newblock Transferable, controllable, and inconspicuous adversarial attacks on
  person re-identification with deep mis-ranking.
\newblock In \emph{CVPR}, 2020{\natexlab{a}}.

\bibitem[Wang et~al.(2020{\natexlab{b}})Wang, Lesner, Ran, Rukonic, Xue, and
  Shiu]{wang2020using}
Wang, W., Lesner, C., Ran, A., Rukonic, M., Xue, J., and Shiu, E.
\newblock Using small business banking data for explainable credit risk
  scoring.
\newblock In \emph{AAAI}, 2020{\natexlab{b}}.

\bibitem[Wang et~al.(2020{\natexlab{c}})Wang, Wang, Ramkumar, Fredrikson,
  Mardziel, and Datta]{wang2020smoothed}
Wang, Z., Wang, H., Ramkumar, S., Fredrikson, M., Mardziel, P., and Datta, A.
\newblock Smoothed geometry for robust attribution.
\newblock \emph{NeurIPS}, 2020{\natexlab{c}}.

\bibitem[Wen et~al.(2020)Wen, Li, and Jia]{wen2020towards}
Wen, Y., Li, S., and Jia, K.
\newblock Towards understanding the regularization of adversarial robustness on
  neural networks.
\newblock In \emph{ICML}, 2020.

\bibitem[Wicker et~al.(2023)Wicker, Heo, Costabello, and
  Weller]{wicker2022robust}
Wicker, M., Heo, J., Costabello, L., and Weller, A.
\newblock Robust explanation constraints for neural networks.
\newblock \emph{ICLR}, 2023.

\bibitem[Wong et~al.(2020)Wong, Rice, and Kolter]{wong2020fast}
Wong, E., Rice, L., and Kolter, J.~Z.
\newblock Fast is better than free: Revisiting adversarial training.
\newblock \emph{ICLR}, 2020.

\bibitem[Xu et~al.(2009)Xu, Caramanis, and Mannor]{xu2009robustness}
Xu, H., Caramanis, C., and Mannor, S.
\newblock Robustness and regularization of support vector machines.
\newblock \emph{JMLR}, 2009.

\bibitem[Xu et~al.(2020)Xu, Ma, Liu, Deb, Liu, Tang, and
  Jain]{xu2020adversarial}
Xu, H., Ma, Y., Liu, H.-C., Deb, D., Liu, H., Tang, J.-L., and Jain, A.~K.
\newblock Adversarial attacks and defenses in images, graphs and text: A
  review.
\newblock \emph{Int. J. Autom. Comput.}, 2020.

\bibitem[Yang et~al.(2020{\natexlab{a}})Yang, Khanna, Yu, Gholami, Keutzer,
  Gonzalez, Ramchandran, and Mahoney]{yang2020boundary}
Yang, Y., Khanna, R., Yu, Y., Gholami, A., Keutzer, K., Gonzalez, J.~E.,
  Ramchandran, K., and Mahoney, M.~W.
\newblock Boundary thickness and robustness in learning models.
\newblock \emph{NeurIPS}, 2020{\natexlab{a}}.

\bibitem[Yang et~al.(2020{\natexlab{b}})Yang, Rashtchian, Zhang, Salakhutdinov,
  and Chaudhuri]{yang2020closer}
Yang, Y.-Y., Rashtchian, C., Zhang, H., Salakhutdinov, R.~R., and Chaudhuri, K.
\newblock A closer look at accuracy vs. robustness.
\newblock \emph{NeurIPS}, 2020{\natexlab{b}}.

\bibitem[Zhang et~al.(2019)Zhang, Yu, Jiao, Xing, El~Ghaoui, and
  Jordan]{zhang2019theoretically}
Zhang, H., Yu, Y., Jiao, J., Xing, E., El~Ghaoui, L., and Jordan, M.
\newblock Theoretically principled trade-off between robustness and accuracy.
\newblock In \emph{ICML}, 2019.

\bibitem[Zhou et~al.(2016)Zhou, Khosla, Lapedriza, Oliva, and
  Torralba]{zhou2016learning}
Zhou, B., Khosla, A., Lapedriza, A., Oliva, A., and Torralba, A.
\newblock Learning deep features for discriminative localization.
\newblock In \emph{CVPR}, 2016.

\bibitem[Zhou et~al.(2020)Zhou, Niu, Wang, Zhang, and Hua]{zhou2020adversarial}
Zhou, M., Niu, Z., Wang, L., Zhang, Q., and Hua, G.
\newblock Adversarial ranking attack and defense.
\newblock In \emph{ECCV}, 2020.

\bibitem[Zhou et~al.(2021)Zhou, Wang, Niu, Zhang, Xu, Zheng, and
  Hua]{zhou2021practical}
Zhou, M., Wang, L., Niu, Z., Zhang, Q., Xu, Y., Zheng, N., and Hua, G.
\newblock Practical relative order attack in deep ranking.
\newblock In \emph{ICCV}, 2021.

\bibitem[Zhou \& Croft(2006)Zhou and Croft]{zhou2006ranking}
Zhou, Y. and Croft, W.~B.
\newblock Ranking robustness: a novel framework to predict query performance.
\newblock In \emph{CIKM}, 2006.

\end{thebibliography}
\bibliographystyle{icml2022}

% \input{checklist}

% \newpage
% \input{supp}

\end{document}